\documentclass[manuscript,noacm=true]{acmart}
\renewcommand\footnotetextcopyrightpermission[1]{}
\usepackage{algorithm}
\usepackage{algpseudocode}
\usepackage{todonotes}
\usepackage{amsmath}
\usepackage{hhline}
\usepackage{subcaption}

\def\BibTeX{{\rm B\kern-.05em{\sc i\kern-.025em b}\kern-.08emT\kern-.1667em\lower.7ex\hbox{E}\kern-.125emX}}
    
\author{Anthony Stein}
\authornote{Corresponding author}
\affiliation{%
  \institution{University of Augsburg}
  \country{Germany}
}
\email{anthony.stein@informatik.uni-augsburg.de}

\author{Roland Maier}
\affiliation{%
  \institution{University of Augsburg}
  \country{Germany}
}
\email{roland.maximilian.maier@student.uni-augsburg.de}

\author{Lukas Rosenbauer}
\affiliation{%
  \institution{BSH Home Appliances}
  \country{Germany}
}
\email{lukas.rosenbauer@bshg.com}

\author{J\"oerg H\"ahner}
\affiliation{%
  \institution{University of Augsburg}
  \country{Germany}
}
\email{joerg.haehner@informatik.uni-augsburg.de}

\title{XCS Classifier System with Experience Replay}

\begin{abstract} 
XCS constitutes the most deeply investigated classifier system today. 
It bears strong potentials and comes with inherent capabilities for mastering a variety of different learning tasks. 
Besides outstanding successes in various classification and regression tasks, XCS also proved very effective in certain multi-step environments from the domain of reinforcement learning.
Especially in the latter domain, recent advances have been mainly driven by algorithms which model their policies based on deep neural networks -- among which the Deep-Q-Network (DQN) is a prominent representative.
Experience Replay (ER) constitutes one of the crucial factors for the DQN's successes, since it facilitates stabilized training of the neural network-based Q-function approximators.
Surprisingly, XCS barely takes advantage of similar mechanisms that leverage stored raw experiences encountered so far. 
To bridge this gap, this paper investigates the benefits of extending XCS with ER.
On the one hand, we demonstrate that for single-step tasks ER bears massive potential for improvements in terms of sample efficiency.
On the shady side, however, we reveal that the use of ER might further aggravate well-studied issues not yet solved for XCS when applied to sequential decision problems demanding for long-action-chains. 
\end{abstract}

\newtheorem{hyp}{Hypothesis}

\begin{document}
\maketitle
\thispagestyle{empty}
\pagestyle{plain}

\renewcommand{\shortauthors}{A. Stein, R. Maier, L. Rosenbauer, J. H\"ahner}

\section{Introduction}
Intriguing advances of Artificial Intelligence (AI) are observable these days. 
One major driver that fueled recent achievements constitutes the paradigm of \textit{Reinforcement Learning} (RL) combined with \textit{Deep Neural Networks} (DNN).
Based on this technology, human-level and even super-human results have been achieved in tremendously complex challenges such as playing Atari games~\cite{mnih2015humanlevel}, defeating the world-best player in Go~\cite{Silver2017} or mastering the real-time strategy game StarCraft II~\cite{Vinyals2019}. 
The \textit{Deep-Q-Network} (DQN)~\cite{mnih2015humanlevel} is a prominent representative of this class of RL algorithms.
It approximates the Q-function by means of utilizing \textit{multi-layer perceptrons} (MLP) which facilitate the learning of very large state spaces in contrast to the `old-fashioned' tabular approach of \textit{Q-Learning}~\cite{Watkins1989}. 
Since non-linear policy approximators such as used in DQN struggle with the inherent online learning nature of RL~\cite{Tsitsiklis1997}, countermeasures against divergences needed to be introduced in order to allow stable learning performance~\cite{mnih2015humanlevel}.
One of those techniques is \emph{Experience Replay} (ER). 
ER alleviates the problem of so-called \textit{detrimental forgetting} by means of breaking the inherent correlations of consecutive samples caused by the online environment interaction of the underlying RL agent. 

The \textit{XCS Classifier System} (XCS) constitutes another learning approach for solving RL problems.
XCS is the most investigated \textit{Learning Classifier System} (LCS) these days, both in terms of empirical evaluation as well as formal theoretical analysis~\cite{Patzel2019}.
XCS and direct descendants proved very successful in a variety of learning tasks, among which classification~\cite{Bernado-MansillaGarrell-Guiu2003,UrbanowiczBertasiusMoore2014}, regression~\cite{ButzLanziWilson2008,SteinMenssenHaehner2018}, data mining~\cite{Kharbat2008,Wilson2001}, and control~\cite{Butz2005c,Bull2004a,SteinRudolphTomfordeEtAl2017} appear most prominent in the literature. 
However, the repeated utilization of previously encountered raw experiences in XCS, as done by DQN via ER, has only barely been subject of research (cf. Sect.~\ref{sec:rw}). 

With a motivation slightly different from that using ER in DQN, this paper proposes and ER-extended XCS variant for real-valued state spaces and examines the potentials as well as remaining challenges of using ER within XCS.

The main scientific contribution of this work is to provide first insights in view of our underlying research hypothesis: 
\begin{hyp}
    XCS with experience replay will profit from already seen raw $(s,a,r,s')$-experiences stored in a limited replay memory in terms of i) an increased sample efficiency, and, ii) improved overall learning capabilities.
\end{hyp}
Further contributions comprise: 
(1) A comparative study of XCS with and without ER and DQN on well-known RL tasks from the OpenAI gym~\cite{openai}. 
(2) A reconsideration and empirical substantiation of an already identified issue regarding the ability of XCS to learn in long-action-chain demanding RL scenarios.

We proceed with first briefly appreciating previous endeavors of using memory mechanisms and stored experiences in the context of LCS in Section~\ref{sec:rw} and then provide necessary background information regarding ER and XCS in Section~\ref{sec:bg}. 
Based on that, we then propose an ER-extended version of XCS in Section~\ref{sec:xcser}.
In order to confirm our hypothesis, in Section~\ref{sec:eval} results from a number of conducted empirical validation studies on single- as well as multi-step problems are reported.
Section~\ref{sec:conclusion} closes the paper with a short summary and a delineation of future work. 

\section{Related Work}\label{sec:rw}
Our literature review did not reveal an explicit use of ER in XCS or directly related systems that are employed in online learning settings. 
Overall, the use of mechanisms to introduce a short- or rather long-term memory through remembering past raw experiences, e.g., state(-action)-reward(-state) tuples which are not further used for inducing a model, has barely attracted interest in the past four decades of LCS research.  
However, it was indeed mentioned to be a viable solution for the overgeneralization issue as identified by Lanzi in~\cite{lanzi1999analysis}.
Stein et al. introduced a so-called \emph{Interpolation Component} to XCS in~\cite{SteinRauhTomfordeEtAl2016,SteinRauhTomfordeEtAl2017}. In their work, experiences are stored as sampling points for later interpolation calculations used to guide the action-selection regime. 
In a later work~\cite{SteinMenssenHaehner2018}, Stein et al. proposed to reuse experienced inputs with their corresponding rewards for modeling the classifier predictions by means of RBF-interpolation in a fuction approximation context.
A memory mechanism was also introduced by Cliff and Ross in~\cite{CliffRoss1994} to allow for learning in non-markovian environments.
Similar aspirations have been followed by Lanzi and Wilson~\cite{Lanzi1998,lanzi2000toward} as well as Pickering and Kovacs~\cite{PickeringKovacs2015} later on.
%
%
Anticipation as an indirect memory mechanism gave rise to another branch of LCS, called \textit{Anticipatory LCS} (ALCS). Representatives comprise ACS~\cite{stolzmann1998anticipatory,butz1999action}, ACS2~\cite{butz2001algorithmic} and YACS~\cite{gerard2002yacs}.
The prototypical scheme of an ALCS can be found in Sutton's \textit{Dyna} architecture~\cite{sutton1991dyna}. 
ALCS constitute instances of \textit{model-based RL} approaches; the model hereby is the internal representation of the environment that is retained in the agent's memory. 

\section{Background}\label{sec:bg}
This section briefly introduces necessary concepts from the domain of RL for the sake of self-containedness. Nevertheless, due to limited space, we need to assume a certain familiarity with temporal difference learning in general as well as DQN and XCS in particular; for a more detailed introduction, the reader is referred to~\cite{sutton1998rli},~\cite{mnih2015humanlevel} and~\cite{wilson1995classifier,butz2002algorithmic}, respectively. 

\subsection{Experience Replay \& Q-Learning}

\emph{Experience Replay} (ER) has been first proposed by Lin in~\cite{lin1992self}. 
The main motivation was to speed up learning of RL agents by introducing a straightforward to implement memory mechanism that facilitates an effective reuse of prior experiences. 
ER is well-recognized as being a possible solution to certain issues present in (deep) RL:
\begin{itemize}
    \item Inherently highly correlated data samples: ER \emph{decorrelates} consecutive samples (i.e., environment observations) which are the consequence of the online interaction with the environment determined by the agent's policy $\pi$. Due to randomized re-sampling of previous experiences from a memory, ER prevents an MLP-based Q-function approximator from known issues such as parameter oscillations or even divergence~\cite{Tsitsiklis1997,mnih2015humanlevel}.  
    \item Detrimental forgetting: In environments where long-action-chains are necessary to reach the desired goal, appropriate credit-assignment of stage-setting transitions is difficult when the underlying Q-function is approximated by neural networks. Under certain circumstances, this can lead to detrimental forgetting which disrupts the sustenance of the action-chain. 
    Given an appropriate replacement and re-sampling strategy, ER allows to remember and favor such \emph{valuable} experiences, e.g., rare transitions that lead to the sparsely available reward.
\end{itemize}
In summary, ER ensures a more decorrelated and thus smoothed training of non-linear models such as (deep) neural networks, leading to (1) a higher \emph{sample efficiency}, and, (2) a more comprehensive approximation of the underlying state-action-payoff landscape. 

For bringing ER to work, first a so-called \emph{replay memory} (RM) has to be added to the learner. 
An RM can be interpreted as a short-to-mid term memory (depending on its size) that stores a certain number of experiences -- in its most simplistic variant this is realized by a \emph{first-in-first-out} (FIFO) queue. 
More formally, let $RM = \langle e_1,...,e_N \rangle$ be a FIFO-buffer of capacity $N$. Each experience $e_i \in RM$ comprises a $(s, a, r, s')$-quadruple.
In each discrete learning step, the RL agent samples an $m$-sized mini-batch of experiences from RM to repeatedly train its model on the memorized experiences. Following its most basic form, as subject of investigation in this paper, the experiences are drawn uniformly at random, i.e., $e_i \sim U(RM)$ 
As already outlined above, an individual experience $e_i \in RM$ is given as a $(s, a, r, s')$-quadruple. 

In the context of temporal difference learning, it is usually assumed that the learning environment is non-deterministic, i.e., characterized by stochasticity. 
Thus, the \textit{succeeding state} $s'$ that a learning agent perceives after executing its \textit{selected action} $a_{exec}$ is not deterministic. 
Accordingly, the probability of the next state can be described by a transition function $P_{s'} (s,a) = P(s' | s,a)$.  
Therefore it follows that the received reward $r$ at a time is also a random variable. 
Let $R(s,a)$ denote the expected reward if the agent is in a current state $s$ of the state space $S$ and decides to take an action $a$ out of the action space $A$. 
With all the ingredients at hand, the resulting quadruple $\langle S, A, P_{s'}(\cdot), R(\cdot)\rangle$ finally defines an MDP for which the learning agent aims to find the optimal policy $\pi^*$.
A policy $\pi:S\rightarrow A$ can be understood as a learning agent's decision mechanism, or as its \emph{action-selection-regime}. It maps the currently observed state $s \in S$ to an action $a_{exec} \in A$ to be executed.  












Q-Learning learns its optimal policy by building a so-called\textit{ state-action value function}, or Q-function. 
In its basic variant, Q-Learning builds up a table comprising Q-value estimates for each possible state-action pair. 
The tabular scheme prohibits the application of Q-learning to large or continuous problem spaces as it is computationally infeasible to maintain highly scaled Q-tables.  
Modern approaches therefore introduce approximators for learning the underlying $Q$-function while facilitating the required generalization. 
The model parameters $\theta$ are trained in a rather supervised manner and the resulting model then maps each state to a corresponding Q-value estimate for each possible action.
Probably the most prominent approach for modern Q-value estimation is the \emph{Deep-Q-Network} (DQN) algorithm~\cite{mnih2013playing} which employs a DNN as the Q-function approximator, called the Q-network.
The parameters $\theta$, i.e., the weights of the Q-network, are trained based on (batch) \emph{stochastic gradient descent} using a quadratic error term as loss function.
However, as already discussed above, certain issues arise when non-linear function approximators, such as MLPs, are utilized to model the Q-function. 
For that reason, ER was introduced to the DQN algorithm, along with further modifications to ensure data efficient and stable learning (for more details cf.~\cite{mnih2013playing,mnih2015humanlevel}). ER allows to re-sample minibatches from the RM which are then used to train the Q-network in sequence or batch-wise using the squared error term (or its expected value in case of batch gradient descent) between the target $r+\gamma \max_{a'}Q(s',a',\theta)$ and the current Q-network's estimate $Q(s,a,\theta)$, effectively stabilizing the training process.

\subsection{XCS Classifier System in a Nutshell}
The XCS classifier system~\cite{wilson1995classifier,wilson1998generalization} is an evolutionary rule-based machine learning system. 
It belongs to the family of \emph{Michigan-style} LCS which are due to Holland~\cite{Holland1978,Holland1986}.\footnote{Our work focuses on Michigan-style LCS mostly characterized by their online learning nature which contrasts the learning intuition of other LCS branches such as Pittsburgh-style~\cite{Smith1980} and Iterative Rule Learning~\cite{BacarditBurkeKrasnogor2009} systems.}
The XCS is an accuracy-based system what distinguishes it from its so-called strength-based relatives such as ZCS~\cite{Wilson1994}. 
Many extensions to the original XCS have been proposed. 
One enhancement of special interest is the introduction of interval predicates along with slight modifications in order to deal with real-valued inputs~\cite{Wilson2000} -- the resulting extension is called \textit{XCS for real valued inputs} (XCSR).\footnote{This work is based on XCSR, but for simplicity we stick to the abbreviation XCS.} 
The next paragraphs very briefly introduce the general learning scheme of XCS. 

At the heart of XCS a population $[P]\coloneq \{cl_i\}_{i\in\mathbb{N}}$ of rules is maintained. Historically, rules are also termed `classifiers'. Here each rule is denoted by $cl_i$ or simply $cl$. A classifier comprises: 
\begin{itemize}
    \item A \textit{condition} $cl.C \subseteq S$ determining a subset of the state space $S$ for which this rule matches
    \item An \textit{action} $cl.a \in A$ from the action space this rule advocates
    \item Quality parameters to be learned online:
        \begin{itemize}
            \item A \textit{payoff prediction} $cl.p$ which can be a simple scalar or computed by a predictive model
            \item An estimate of the absolute \textit{prediction error} $cl.\epsilon$ 
            \item A rule's \textit{fitness} $cl.F$, estimating its niche-relative payoff prediction accuracy
        \end{itemize}
    \item Further book-keeping statistics:
            \begin{itemize}
                \item $cl.exp$ denotes the number of conducted reinforcements  
                \item $cl.ts$ is a time-stamp updated every time this classifier was a candidate for the GA  
                \item $cl.as$ estimates the average size of action sets this classifier has been part of
                \item $cl.num$ counts successful subsumption operations
        \end{itemize}
\end{itemize}

XCS's working principle can be described as the result of a complex interaction between several components: \textit{Local learning} of the payoff prediction models, \textit{global optimization} of the classifiers' locality (i.e., the size of their conditions $cl.C$), \textit{collective decision making} based on a fitness-weighted mixing of the matching rules' predictions, as well as \textit{population-wide replacement} (i.e., insertion and deletion) mechanisms to satisfy the population size constraint $\sum_{cl\in[P]}cl.num\leq N$. The \textit{numerosity} $cl.num$ is a counter of successful subsumptions which further facilitates \textit{generalization} over the state space~\cite{wilson1998generalization}.
In standard XCS, local learning is conducted by employing a \textit{Widrow-Hoff} learning scheme to the quality parameters
of rules matching the current state $s$ (hold in a so-called \textit{match set} $[M]\subseteq [P]$) and whose action was selected for being executed (hold in another subset called the \textit{action set} $[A]\subseteq [M]$).  
A \textit{steady-state niche Genetic Algorithm} (GA) in combination with a population-wide (panmictic) classifier replacement mechanism takes care of global optimization toward maximally general and accurate rules. 
The inference or \textit{action-selection} step greedily selects the highest fitness-weighted prediction sum of all classifiers in $[M]$ that vote for the same action. 
Therefor a \textit{system prediction} for each action present in $[M]$ is stored in the \textit{prediction array} (PA). 
In order to account for the exploration/exploitation dilemma, actions are drawn by chance with a certain probability (e.g., using $\epsilon$-greedy policy) as usual for model-free RL agents. 

XCS is very flexible in terms of the learning tasks it is able to solve -- thereby only demanding for a few straightforward modifications. 
XCS is not restricted to just RL problems (single- or multi-step) but has also been very successfully applied to supervised classification and regression tasks~\cite{Bernado-MansillaGarrell-Guiu2003,UrbanowiczBertasiusMoore2014,ButzLanziWilson2008,SteinMenssenHaehner2018}, data mining~\cite{Wilson2001,Kharbat2008} as well as unsupervised clustering~\cite{TameeBullPinngern2007} and even autoencoding~\cite{preen2019autoencoding}.

Algorithm~\ref{algo:main_loop} provides a concise description of one iteration through XCS's main learning loop. This loop is passed through as long as the selected termination criteria are not met, e.g., a maximum number of learning steps or convergence.  

\begin{algorithm}[!htpb]
\caption{XCS learning iteration, adapted from~\cite{butz2002algorithmic}}
\label{algo:main_loop}
\begin{algorithmic}[1]
\State 	Observe environmental state $s_t$ at time $t$
\State 	Create match set $[M] \subseteq [P]$ for $s_t$, i.e., $[M]\coloneq \{cl_i | s_t \in cl_i.C\}$
\State 	Create $PA$ from $[M]$, i.e., $PA(a)=\frac{\sum_{cl \in [M]|cl.a = a}cl.p\cdot cl.F}{\sum_{cl \in [M]|cl.a = a} cl.F}$
\If     {$U[0,1] < \epsilon$} 
\State  Select random action, i.e., $a_{exec}=rand(A)$
\Else
\State 	Select best action, i.e., $a_{exec}=\text{argmax}_{a}PA(a)$
\EndIf
\State 	Create action set $[A] \subseteq [M]$, i.e., $[A]\coloneq \{cl_i | cl_i.a = a_{exec}\}$ 
\State  Execute $a_{exec}$ on the environment
\State  Observe environmental reward $r_t$
\If {$[A]_{t-1}\neq \emptyset$}
\State $P \leftarrow r_{t-1} + \gamma \cdot \text{argmax}_{a}PA(a)$
\State Update all $cl \in [A]_{t-1}$ using $P$
\If {$t-\frac{\sum_{cl\in[A]_{t-1}} cl.num\cdot cl.ts}{\sum_{cl\in[A]_{t-1}} cl.num} > \theta_{GA}$}
\State Run GA on $[A]_{t-1}$ considering $s_{t-1}$
\EndIf
\EndIf
\If {end of episode}
\State $P \leftarrow r_t$
\State Update all $cl \in [A]$ using $P$
\If {$t-\frac{\sum_{cl\in[A]} cl.num\cdot cl.ts}{\sum_{cl\in[A]} cl.num} > \theta_{GA}$}
    \State Run GA on $[A]$ considering $s_t$
\EndIf
\State $[A]_{t-1} \leftarrow \emptyset$
\Else 
\State $[A]_{t-1} \leftarrow [A]$
\State $r_{t-1} \leftarrow r_t$
\State $s_{t-1} \leftarrow s_t$
\EndIf
\end{algorithmic}
\end{algorithm}

\section{XCS with Experience replay}\label{sec:xcser}
One important aspect to note is that the credit assignment step in XCS shares obvious similarities to the Q-value update~\cite{wilson1995classifier,Lanzi2002}. 
The most distinguishing aspect, however, is the GA-based state-space generalization capability of XCS compared to tabular Q-learning. 
The rules' conditions $cl.C$ essentially partition the state space $S$ which leads to a local learning intuition. 
This means that not one global approximation of the state-action-payoff map $S \times A \rightarrow P$ is learnt at once.
Instead, the problem space is partitioned and individual niches are approximated by several, partially overlapping classifiers, following the divide-and-conquer principle. 
The actual partitioning is optimized by the employed GA. Its selection pressure strongly depends on the niche-relative accuracy of the candidate rules ($cl \in [A]$). 
A rule's accuracy follows a sort of inverse relationship to its prediction error estimate $cl.\epsilon$, more precisely:
%
\begin{equation}
    cl.\kappa = 
    \begin{cases}
    1 & ,~cl.\epsilon < \epsilon_0\\
    \alpha \bigl(\frac{\epsilon}{\epsilon_0}\bigr)^{-\nu} & ,~\text{otherwise}
    \end{cases}    
\end{equation}
From that it follows, that the appropriate generalization toward maximally general and at the same time maximally accurate rules depends strongly on the \textit{data efficiency} of the employed learning mechanisms; in case of standard XCS the Widrow-Hoff, also known as least mean squares method. 
This in turn implies that an increased convergence rate of the classifiers' error estimates $cl.\epsilon$ -- e.g., due to ER -- results in faster perceptibility of an appropriate fitness signal fueling the GA during classifier generalization.

For that reason, we propose to leverage the potentials of ER regarding its ability to increase data efficiency by repeatedly feeding the learning rules with memorized experiences.
Therefor, we modified the main learning loop of XCS as given in Algorithm~\ref{algo:xcs-er}:

\begin{algorithm}[htpb!]
\caption{XCS with ER}\label{algo:xcs-er}
\begin{algorithmic}[1]

\State Initialize $RM$ with capacity $|RM|=N$ as FIFO-queue

\Repeat 

    \State  Follow \textbf{Alg.~\ref{algo:main_loop}} until \textbf{line 11}
    \Comment{see \textbf{Alg.~\ref{algo:main_loop}}}
    \Statex \textit{// Add experience to $RM$}
    
    \If {$[A]_{t-1}\neq \emptyset$}
        \Comment{Multi-step problem}
        \State $RM \coloneq RM \cup \{(s_{t-1}, a_{t-1}, r_{t-1}, s_t)\}$
        \State $s_{t-1} \leftarrow s_t$
        \State $a_{t-1} \leftarrow a_{exec}$
        \State $r_{t-1} \leftarrow r_t$
        \State $[A]_{t-1} \leftarrow [A]$
        \Else
        \Comment{Single-step problem}
        \State $RM \coloneq RM \cup \{(s_{t}, a_{exec}, r_{t})\}$
    \EndIf 
    
    \State Possibly drop oldest experience
    \Comment{FIFO-queue}
    \Statex \textit{// Sample minibatch of size $m$ from $RM$}
    \State $B_{exp} \coloneq \{e_j | e_j \in RM \land e_j \sim U(RM)\}_{j=1\dots m}$
    
    \For{$e_j : B_{exp}$}
        \State Do \textsc{ExperienceReplay}($e_j$)
        \Comment{see \textbf{Alg.~\ref{algo:er-reinforcement}}}
    \EndFor
    
    \If{end of episode}
        \Comment{Always true for single-step}
        \State $[A]_{t-1}\leftarrow \emptyset$
    \EndIf

\Until {termination criteria met}

\end{algorithmic}
\end{algorithm}

\begin{algorithm}[htpb!]
\caption{ER reinforcement}\label{algo:er-reinforcement}
\begin{algorithmic}[1]
\Procedure{ExperienceReplay}{$e_j$}
\State  Read $e_j\coloneq(s, a_{exec}, r, s')$ quadruple
\State 	Create match set $[M] \coloneq \{cl | s \in cl.C \land cl \in [P]\}$
\State 	Create action set $[A] \coloneq \{cl | cl.a = a_{exec} \land cl \in [M]\}$ 

\If {$s'$ is not terminal}
    \State 	Create match set $[M]' \coloneq \{cl | s' \in cl.C \land cl \in [P]\}$
    \State 	Create $PA$ with $PA(a)=\frac{\sum_{cl \in [M]'|cl.a = a}cl.p\cdot cl.F}{\sum_{cl \in [M]'|cl.a = a} cl.F}$
    \State $P \leftarrow r + \gamma \cdot \text{argmax}_{a}PA(a)$
\ElsIf {$s'$ is terminal $|$ not existent} 
    \Comment{Single-step}
    \State $P \leftarrow r$
\EndIf

\Statex // \textit{Update all} $cl\in [A]$ \textit{using} $P$
\For{$cl : [A]$}
    \State $cl.exp \leftarrow cl.exp + 1$
    \State $cl.\epsilon \leftarrow cl.\epsilon + \beta\cdot (|P - cl.p|-cl.\epsilon)$
    \State $cl.p \leftarrow cl.p + \beta\cdot (P-cl.p)$
    \State $cl.F \leftarrow cl.F + \beta\cdot (cl.\kappa'-cl.F)$
    \State $cl.as \leftarrow cl.as + \beta\cdot (\sum_{cl \in [A]}cl.num-cl.as)$
\EndFor
    \If {$t-\frac{\sum_{cl\in[A]} cl.num\cdot cl.ts}{\sum_{cl\in[A]} cl.num} > \theta_{GA}$}
        \State Run GA on $[A]$ considering $s$
    \EndIf
\EndProcedure
\end{algorithmic}
\end{algorithm}

As can be seen from Algorithms~\ref{algo:xcs-er} and~\ref{algo:er-reinforcement}, mainly the credit assignment
and the GA application frequency have been modified. 
Instead of the conventional one-pass learning approach which is based on the most recently seen experience $e_t=(s_{t-1},a_{t-1},r_{t-1},s_t)$ or $e_t=(s_{t},a_{exec},r_{t})$ for multi- and single-step problems, respectively, a predefined number of previous experiences $e_j \in RM$ is used.
The current experience is only added to the limited replay memory $RM$ which is realized as a FIFO-buffer.
In its simple form, as introduced in this paper, all experiences stored in $RM$ get the same chance of being selected. 
This essentially ensures that the underlying dynamics in terms of the sample distribution of the learning problem is not affected, as would be the case when more sophisticated concepts such as \textit{prioritized replay}~\cite{schaul2015prioritized} would be employed (which is not subject of this initial work). 
The actual learning then happens on the basis of a minibatch $B_{exp}$ of predefined size $m$ which is sampled uniformly at random from the $RM$. 
ER reinforcement (cf. Alg.~\ref{algo:er-reinforcement}) is then executed for each experience $e_j \in B_{exp}\subseteq RM$. 
Depending on whether XCS is asked to solve a single- or multi-step (sequential) problem, different \textit{payoff values} $P$ are considered for the credit assignment or update steps as usual.
For each experience $e_j \in B_{exp}$ which is going to be replayed, the employed steady-state niche GA gets a chance of being invoked. This effectively increases the GA frequency by a factor $m$, i.e., the number of experiences drawn from $RM$ for the minibatch.
Increased GA application exerts higher evolutionary pressure to the population $[P]$, more precisely to the niches which are hit by the drawn replay experiences. 

In a fully uniformly sampled state space, ER in XCS is expected to directly increase the convergence speed.  
As we will show in the subsequent evaluation section, for problems subject to non-uniform problem space sampling, the above assumption might not hold. 
Tasks belonging to the latter category are almost always given in sequential problems (recall the above discussion of the highly correlated consecutive samples). 
In a multi-step scenario, the application of ER was expected to support the learning progress especially in the presence of sparse reward signals. 
But as we will demonstrate below, ER at the same time bears the risk of boosting the well-recognized issue of premature overgeneralization~\cite{lanzi1999analysis,Barry2002}, which hinders XCS from sustaining long-action-chains.   
In the following, XCS using ER is denoted as XCS-ER. 
The capacity of the RM and the size of the resampled mini-batches for conducting ER was fixed to $N=50k$ $m=4$ in this work, respectively. Furthermore, we applied a warm-up phase over the first 1000 learning steps to fill the RM. Learning only starts after this warm-up.  




\section{Evaluation}\label{sec:eval}
This section summarizes the results we obtained on a variety of experiments conducted in order to assess the potentials of XCS using ER. 
We conducted experiments on three well-known single-step problems (see Sect.~\ref{sec:singlestep}) as well as on three challenging tasks from the domain of multi-step RL (see Sect.~\ref{sec:multistep}).
All conducted experiments have been repeated 30 times with different random seeds. 
The tables below (Tab.~\ref{tab:single-step}) and in the supplemental material indicate the repetition means as well as the observed standard deviations.
Standard XCS without ER serves as baseline for statistical assessment. 
First, we conducted \textit{Shapiro-Wilk tests} to check for normal distribution of the repetition means. 
In the positive case, \textit{paired one-sided t-tests} between the baseline and XCS-ER (and additionally DQN for multi-step) have been selected to judge statistical significance of the differences. 
Otherwise, the \textit{Wilcoxon signed-rank test} was conducted. 
During the multi-step experiments, a DQN baseline implementation as provided by OpenAI~\cite{openai} served as further opponent. 
For comparability reasons, vanilla DQN as proposed by Mnih et al.~\cite{mnih2015humanlevel} with standard ER has been chosen. 
The plots as depicted below show the learning curves of the compared algorithms over the entire learning period. Error bars indicate the standard deviation. 
For the single-step experiments, the learning progress is depicted in terms of the aggregated reward as well as the system prediction error as commonly done in the LCS literature. 
To evaluate the progress in the multi-step scenarios, the last 100-episode reward mean is plotted as often done in the RL field. 
The axis scales depend on the reward schemes of the problem under investigation.

\subsection{Results on Single-step Problems}\label{sec:singlestep}
First, XCS-ER was evaluated on three benchmark problems well-known in the context of LCS-research. 
These are: (1) The \emph{real-multiplexer problem} (RMP)~\cite{Wilson2000,StoneBull2003}, (2) the \textit{Mario} classification problem as introduced in~\cite{SteinMaierHaehner2017}, and, (3) the \textit{Wisconsin Breast Cancer} (WBC) data set from the domain of medical data mining~\cite{Kharbat2008,Wilson2001}.
The RMP is a challenging problem for machine learning systems, since it exhibits characteristics such as heterogeneity (multiple niches with equal actions) and epistasis (feature interactions)~\cite{UrbanowiczMoore2015}. RMP is a real-valued binary classification problem. In this paper, we used the $k=6$ RMP variant. 
The Mario problem is a generalization of the checkerboard problem~\cite{StoneBull2003}, a well-known benchmark for assessing LCS. 
It comprises a 16x16 pixel-art of Mario, which entails seven possible actions (the different colors) and allows for different generalizations in different niches (e.g., the blue trousers in contrast to the yellow knobs). Mario constitutes a two-dimensional real-valued multi-class classification problem.
The WBC data set\footnote{https://archive.ics.uci.edu/ml/datasets/breast+cancer+wisconsin+(original) (07.02.20)} from the UCI repository~\cite{Lichman2013} is probably one of the most famous benchmarks for classification. It consists of 10 integer-valued input dimensions indicating several features from a breast cancer diagnostics assessment and a binary label for `malignant' or `benign' . 
Usually, other types of modern LCS would be preferred for mining such data, e.g., ExSTRaCS~\cite{UrbanowiczBertasiusMoore2014,UrbanowiczMoore2015} or BioHEL~\cite{BacarditBurkeKrasnogor2009}. However, we selected this data set, since (1) XCS was already applied on it and therefore we could make use of tuned hyperparameter-settings, and, (2) we pursue the aim to demonstrate that ER can improve XCS's ability for stream mining of real-world data. 
Furthermore, the presented ER extension appears to be straightforwardly adaptable for any supervised LCS directly descending from XCS.  

For each of the single-step problems, a binary reward scheme has been applied. For each correctly proposed action a reward of 1000 was payed. Otherwise, 0 was returned.
The unordered bound hyperrectangular condition representation has been chosen. Further hyperparameters have been adopted from the literature and can be found in~\cite{Wilson2000,StoneBull2003},~\cite{SteinMaierHaehner2017} and~\cite{Wilson2001,Kharbat2008} for RMP, Mario and WBC, respectively, as well as in this paper's supplemental material. 

Table~\ref{tab:single-step} provides the observed results. 
For the single-step tasks we assessed XCS for four standard figures of merit: (1) The average reward received at the end of the learning period. (2) The average system error between the predicted and the actually received reward. (3) The average size of the population, denoted $|[P]|$. And finally, (4) the average generality of the classifiers in $[P]$, which is equivalent to the average volume of the classifier conditions.

\begin{table}[htbp]
\caption{Overall results for the single-step environments. Reward, system error, population size and generality is indicated. All entries show the means $\pm$1SD over the 30 repetitions. Arrows indicate whether the metric increased or decreased compared to standard XCS. * (**) indicate statistically significant differences, i.e., a p-value < $\alpha$ = 0.05 (0.01). Bold entries highlight improvements over the baseline.}
\begin{tabular}{cllll}
\bottomrule
\textbf{Mario} & Reward & Sys. Err. & $|[P]|$ & Generality \\ 
\toprule
\multicolumn{1}{r}{XCS-ER-8} & \textbf{\begin{tabular}[c]{@{}l@{}}959.67$\uparrow$**\\ $\pm$1.72\end{tabular}} & \textbf{\begin{tabular}[c]{@{}l@{}}49.17$\downarrow$**\\ $\pm$1.90\end{tabular}} & \textbf{\begin{tabular}[c]{@{}l@{}}2151.76$\downarrow$**\\ $\pm$19.5\end{tabular}} & \textbf{\begin{tabular}[c]{@{}l@{}}0.0078$\downarrow$**\\ $\pm$0.0002\end{tabular}} \\ 
\hline
\multicolumn{1}{r}{XCS-ER-4} & \textbf{\begin{tabular}[c]{@{}l@{}}944.60$\uparrow$**\\ $\pm$2.69\end{tabular}} & \textbf{\begin{tabular}[c]{@{}l@{}}69.58$\downarrow$**\\ $\pm$2.94\end{tabular}} & \textbf{\begin{tabular}[c]{@{}l@{}}2314.33$\downarrow$**\\ $\pm$16.39\end{tabular}} & \textbf{\begin{tabular}[c]{@{}l@{}}0.0085$\downarrow$**\\ $\pm$0.0002\end{tabular}} \\ 
\hhline{=====}
\multicolumn{1}{r}{XCS} & \begin{tabular}[c]{@{}l@{}}880.12\\ $\pm$3.54\end{tabular} & \begin{tabular}[c]{@{}l@{}}188.81\\ $\pm$3.70\end{tabular} & \begin{tabular}[c]{@{}l@{}}2759.87\\ $\pm$28.47\end{tabular} & \begin{tabular}[c]{@{}l@{}}0.0092\\ $\pm$0.0002\end{tabular} \\ 
\bottomrule
\textbf{6-RMP} & Reward & Sys. Err. & $|[P]|$ & Generality \\ 
\toprule
\multicolumn{1}{r}{XCS-ER} & \textbf{\begin{tabular}[c]{@{}l@{}}957.64$\uparrow$**\\ $\pm$3.10\end{tabular}} & \textbf{\begin{tabular}[c]{@{}l@{}}67.18$\downarrow$**\\ $\pm$5.89\end{tabular}} & \textbf{\begin{tabular}[c]{@{}l@{}}281.37$\downarrow$**\\ $\pm$11.58\end{tabular}} & \textbf{\begin{tabular}[c]{@{}l@{}}0.0010$\uparrow$**\\ $\pm$4.96E-5\end{tabular}} \\ 
\hhline{=====}
\multicolumn{1}{r}{XCS} & \begin{tabular}[c]{@{}l@{}}925.97\\ $\pm$12.82\end{tabular} & \begin{tabular}[c]{@{}l@{}}147.37\\ $\pm$18.91\end{tabular} & \begin{tabular}[c]{@{}l@{}}409.65\\ $\pm$20.56\end{tabular} & \begin{tabular}[c]{@{}l@{}}0.0008\\ $\pm$5.28E-5\end{tabular} \\ 
\bottomrule
\textbf{WBC} & Reward & Sys. Err. & $|[P]|$ & Generality \\ 
\toprule
\multicolumn{1}{r}{XCS-ER} & \begin{tabular}[c]{@{}l@{}}989.34$\downarrow$**\\ $\pm$0.72\end{tabular} & \textbf{\begin{tabular}[c]{@{}l@{}}15.83$\downarrow$**\\ $\pm$0.77\end{tabular}} & \textbf{\begin{tabular}[c]{@{}l@{}}3714.27$\downarrow$**\\ $\pm$27.20\end{tabular}} & \textbf{\begin{tabular}[c]{@{}l@{}}3.46E-7$\uparrow$**\\ $\pm$ 1.35E-7\end{tabular}} \\ \hhline{=====}
\multicolumn{1}{r}{XCS} & \begin{tabular}[c]{@{}l@{}}996.42\\ $\pm$0.76\end{tabular} & \begin{tabular}[c]{@{}l@{}}24.01\\ $\pm$1.82\end{tabular} & \begin{tabular}[c]{@{}l@{}}3942.97\\ $\pm$26.24\end{tabular} & \begin{tabular}[c]{@{}l@{}}1.00E-7\\ $\pm$2.86E-8\end{tabular}
\end{tabular}
\label{tab:single-step}
\end{table}

As can be noted from Table~\ref{tab:single-step}, XCS-ER significantly outperforms standard XCS without ER in nearly all figures of merit for any of the three examined problems. 
For the Mario pixel art problem, we exemplarily doubled the size of the minibatch to $m=8$ in order to fathom whether more extensive replay bears higher potential for improvements. Although the one conducted preliminary experiment promises further improvements, this aspect needs to be investigated further.
Considering the generality metric, it becomes apparent that for the Mario environment, significantly smaller average generalities have been observed (for detailed plots see supplemental material). However, if we judge this result in conjunction with the observed progress of the population size, it appears that subsumption can act more properly which results in a significantly decreased average number of rules in the population. Paired with the highly improved reward and system error metrics, we conclude that ER has supported XCS to reach a more appropriate degree of rule specification much quicker, thereby increasing sample efficiency as hypothesized at the beginning.  

\begin{figure}[htpb]
\centering
\begin{subfigure}[]{\textwidth}
    \centering
    \includegraphics[scale=0.65]{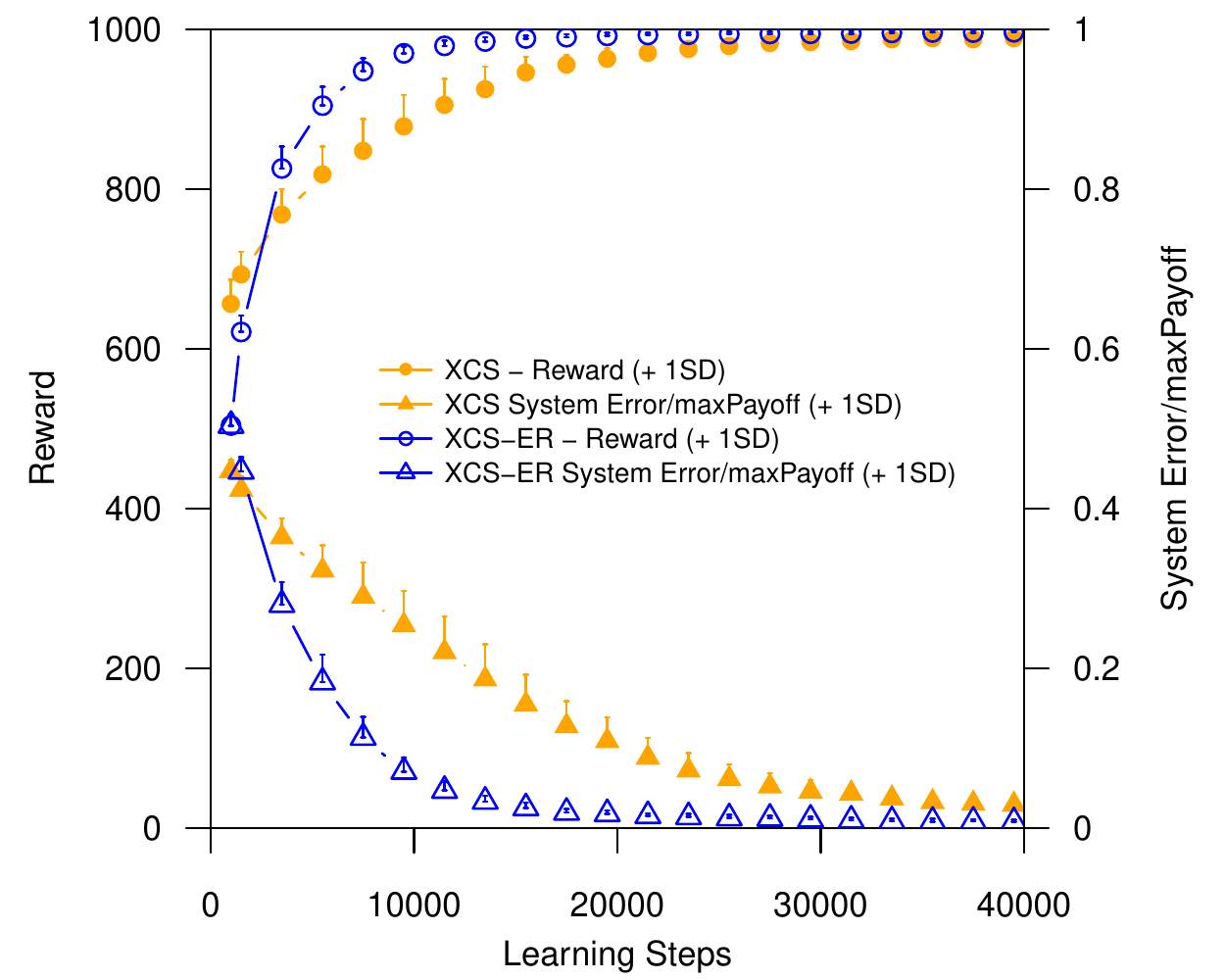}
    \caption{$k=6$-Real Multiplexer}
\end{subfigure}

\begin{subfigure}[]{\columnwidth}
    \centering
    \includegraphics[scale=0.65]{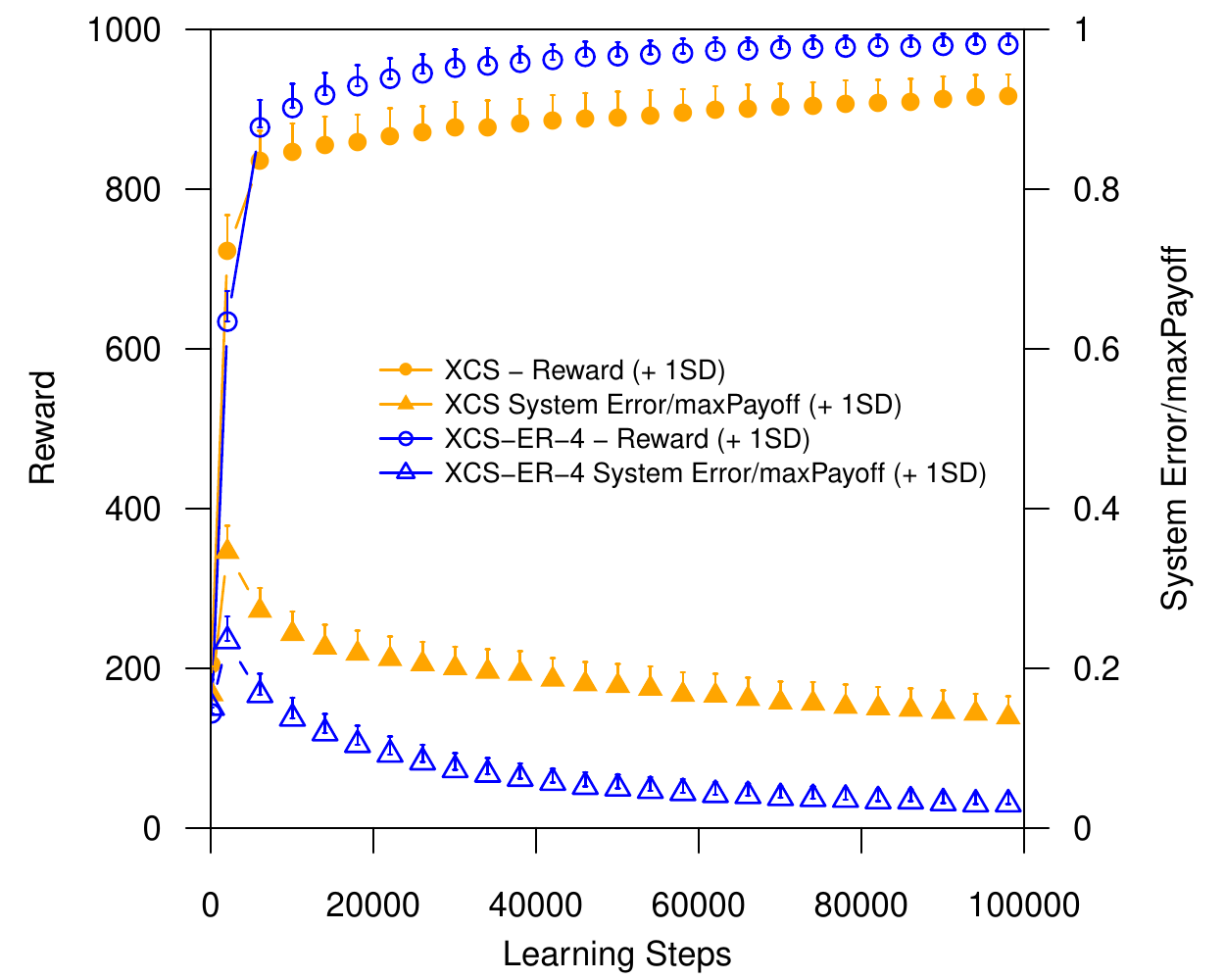}
    \caption{Mario Pixel Art~\cite{SteinMaierHaehner2017}}
\end{subfigure}

\begin{subfigure}[]{\columnwidth}
    \centering
    \includegraphics[scale=0.65]{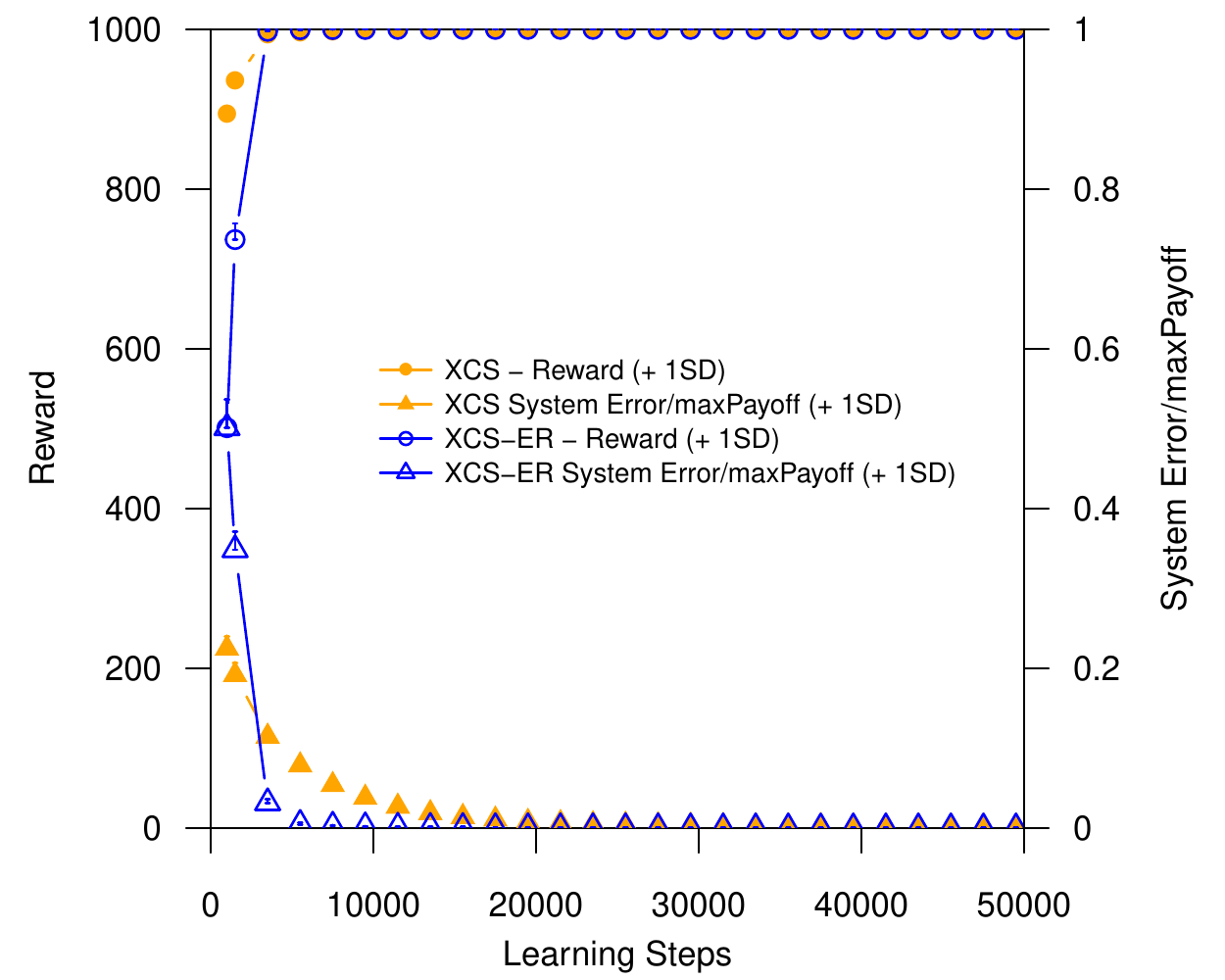}
    \caption{WBC classification}
\end{subfigure}

\caption{XCS vs. XCS-ER on single-step problems.}
\label{fig:eval_singlestep}
\end{figure}

Although the overall performance of XCS on single-step problems can be significantly improved through ER, inspection of the learning plots (Fig.~\ref{fig:eval_singlestep}) reveals one interesting insight. When starting off tabula rasa, i.e., with an empty ruleset $[P]$, during the very initial learning phases XCS-ER appears to have slightly inferior performance in terms of reward and system error. This is attributed to: (1) The initial 1000 steps comprising a warm-up phase where no learning happens. (2) The higher evolutionary pressure exerted through the use of ER. The GA has $m$-times more opportunities to act, which results in more transient rules due to a poor fitness signal resulting in more random variations. This shortly blows up the population at the beginning.  
However, the more frequently applied GA also increases the generalization pressure, since at the same time the error estimates converge more quickly what effectively leads to a stronger fitness signal. The additional plots as given in the supplemental material corroborate this first conjecture, which yet needs more research.  
Especially for the WBC data set, of which already standard XCS is very capable, the abovementioned effect resulted in a slightly but indeed significant decrease of the average reward received over the entire learning period, even if the prediction error itself can be significantly improved.
In order to attenuate this effect, recently proposed interpolation-based action-selection and classifier generation schemes~\cite{SteinEymullerRauhEtAl2016,SteinRauhTomfordeEtAl2017}, found to effectively decrease initial prediction errors, could be employed.   

It can be concluded that XCS for real-valued inputs using ER substantially increases the learning efficiency in single-step problems. 

\subsection{Results on Multi-step Problems}\label{sec:multistep}
XCS has been found to reliably solve certain grid-like multi-step problems. 
However, literature evaluating XCS on well-known RL problems such as CartPole or MountainCar remains sparse. 
We used the OpenAI gym~\cite{openai} to compare XCS and XCS-ER against standard DQN. 
Three challenging environments serve as testbeds. 
(1) The \textit{nChain} problem of length $n=16$. 
(2) The \textit{pole balancing} cart control problem (CartPole). 
(3) The \textit{MountainCar} environment. 
Due to lack of space, we refer the reader to~\cite{sutton1998rli,Strens2000} and the OpenAI gym documentation~\cite{openai} for a detailed description of the tasks. Slight modifications to the environments had to be carried out which do not considerably change the problems' complexities. Please confer the supplemental material for details regarding the modifications and the applied hyperparameter configurations. 

\begin{figure*}[htpb]
\centering
\begin{subfigure}[b]{.47\textwidth}
  \centering
  \includegraphics[width=\textwidth]{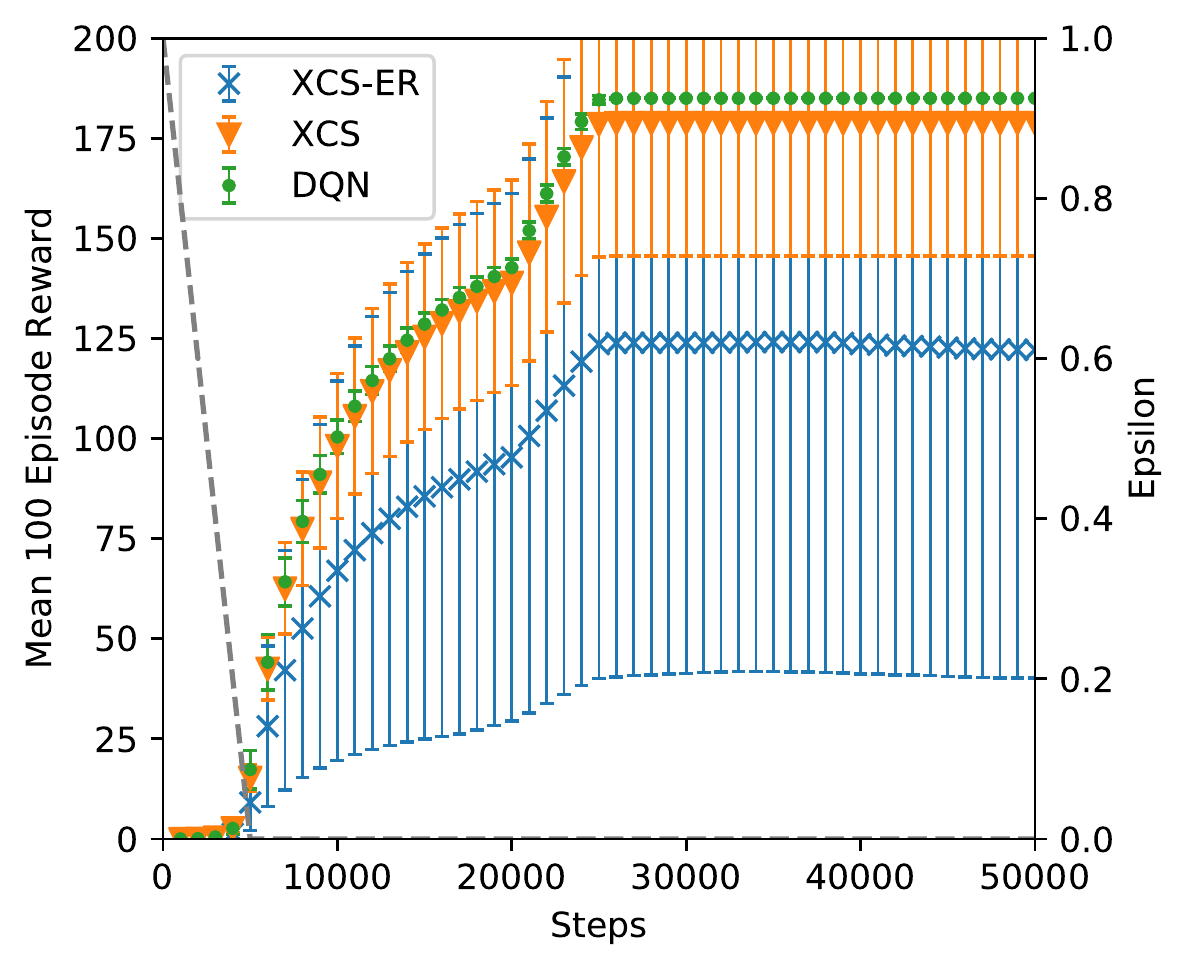}
  \caption{XCS vs. XCS-ER in 16Chain.}
 \end{subfigure}  
\qquad
\begin{subfigure}[b]{.47\textwidth}
  \centering
  \includegraphics[width=\textwidth]{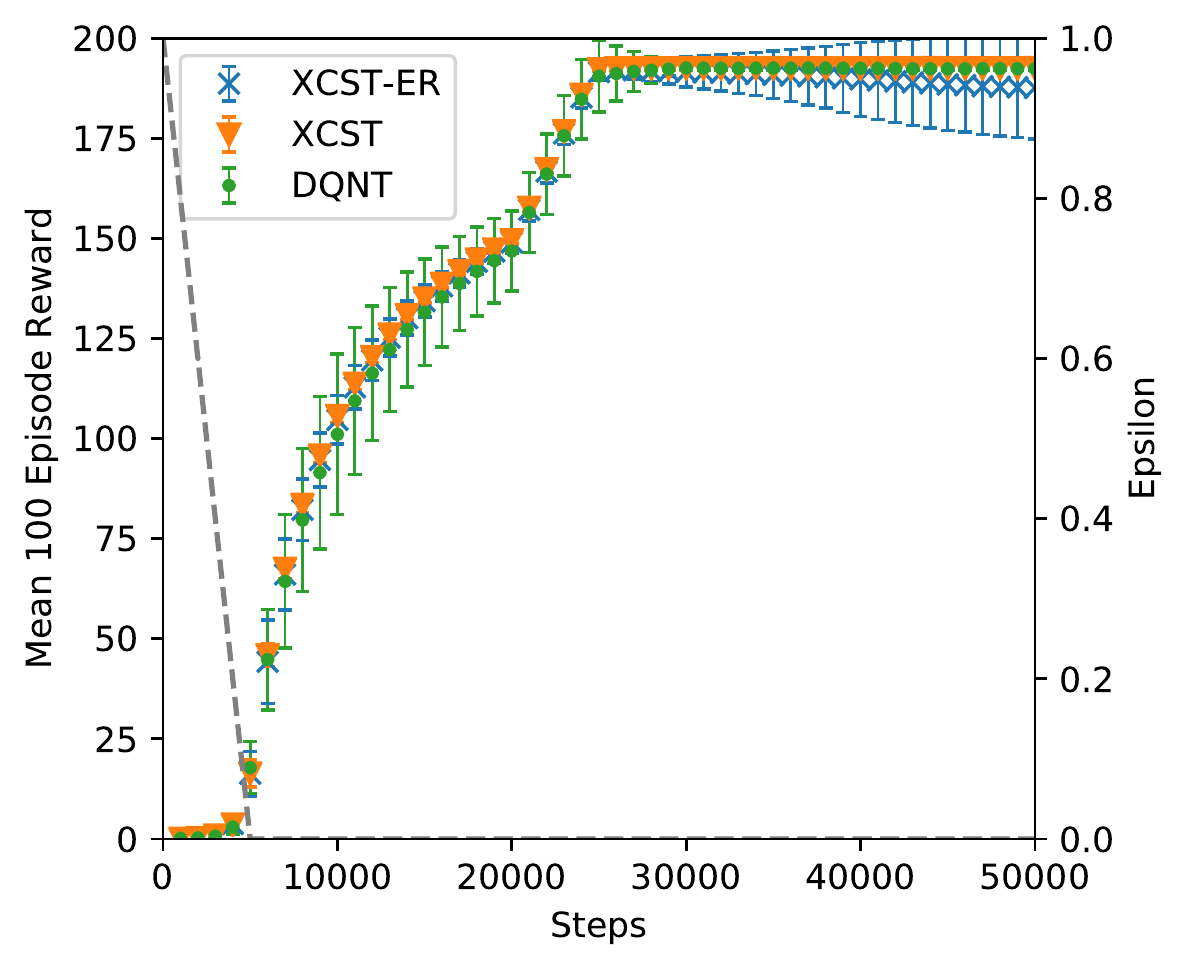}
  \caption{XCST vs. XCST-ER in 16Chain.}
\end{subfigure} 

\begin{subfigure}[b]{.47\textwidth}
    \centering
    \includegraphics[width=\textwidth]{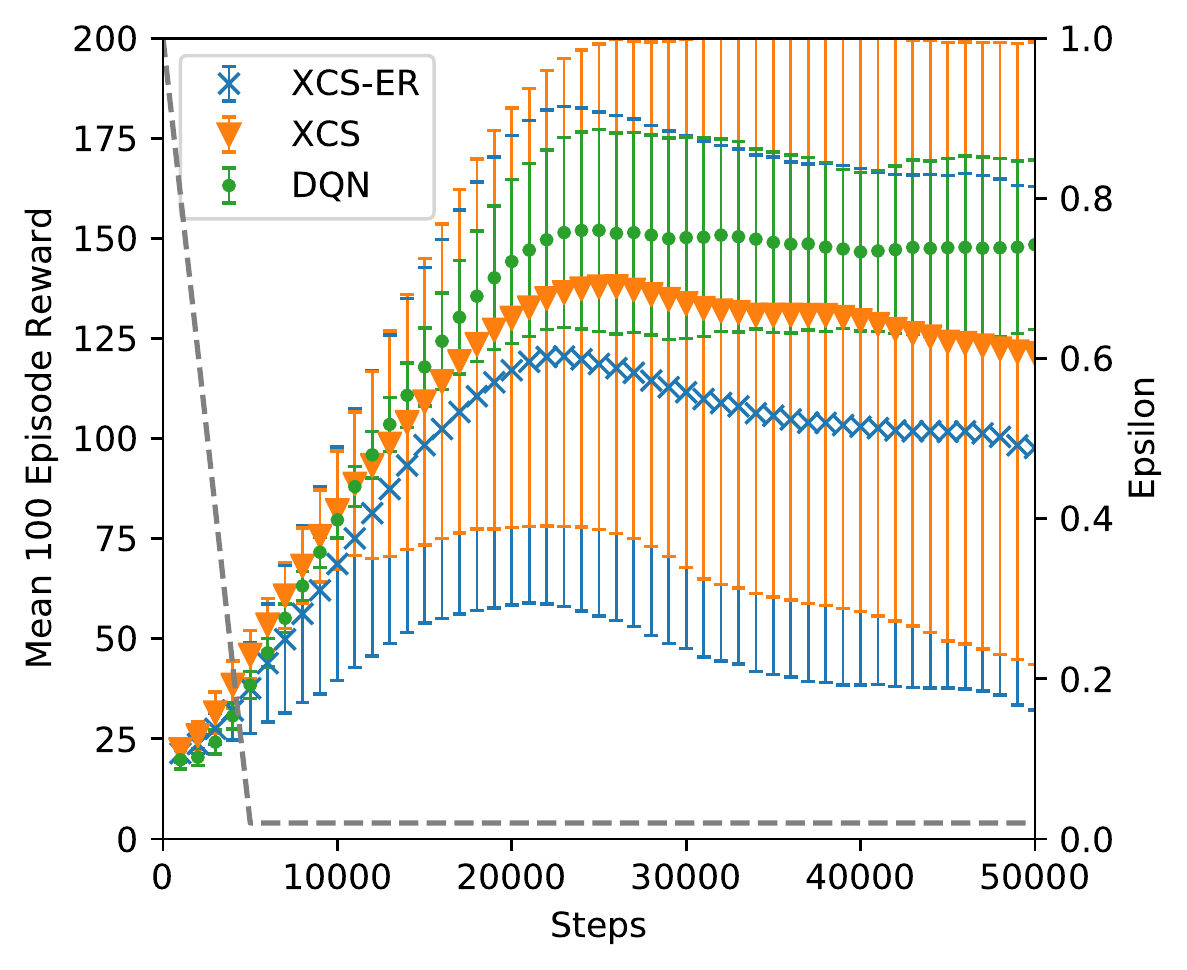}
    \caption{XCS vs. XCS-ER in CartPole.}
\end{subfigure}  
\qquad
\begin{subfigure}[b]{.47\textwidth}
    \centering
    \includegraphics[width=\textwidth]{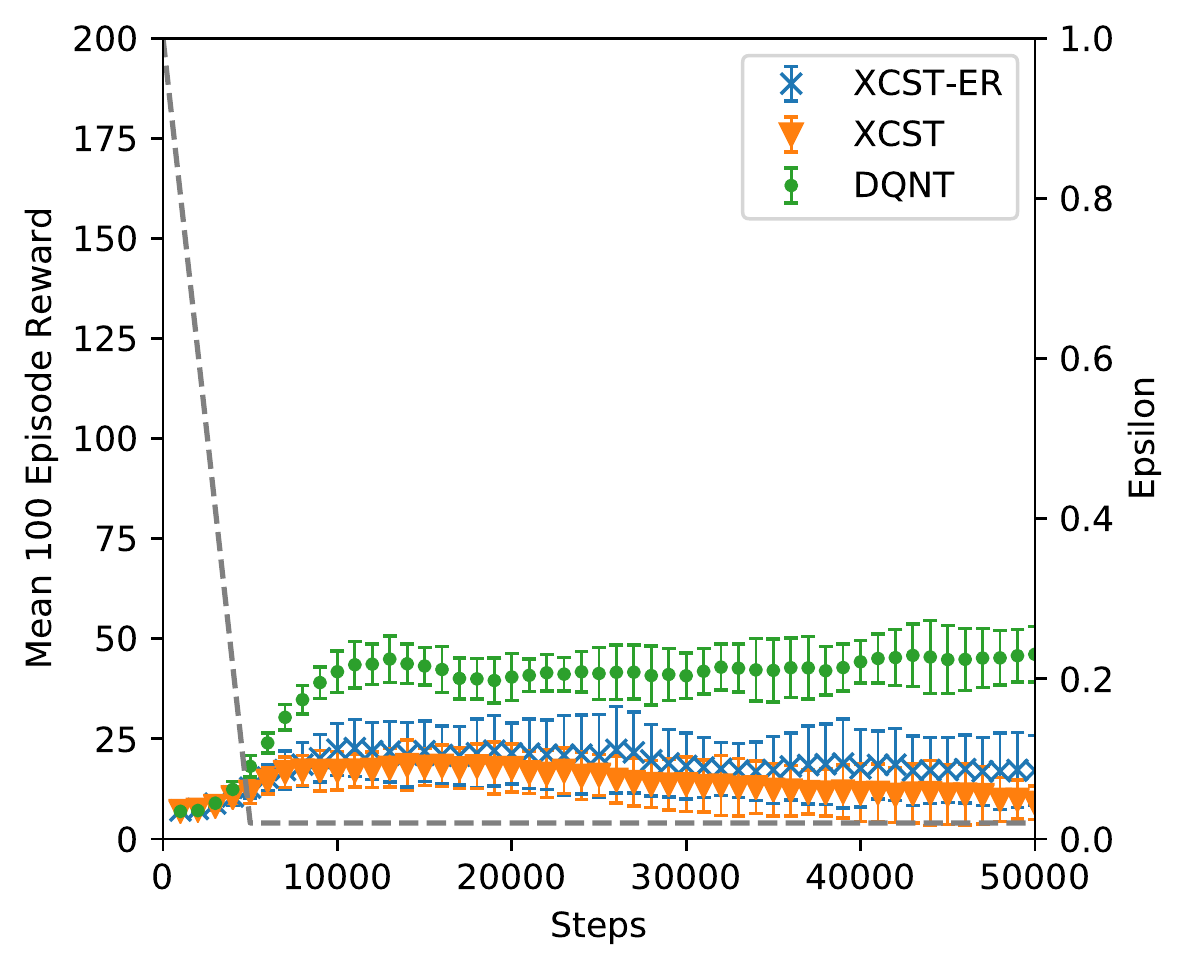}
    \caption{XCST vs. XCST-ER in CartPole.}
\end{subfigure}  
    
\begin{subfigure}[b]{.47\textwidth}
    \centering
    \includegraphics[width=\textwidth]{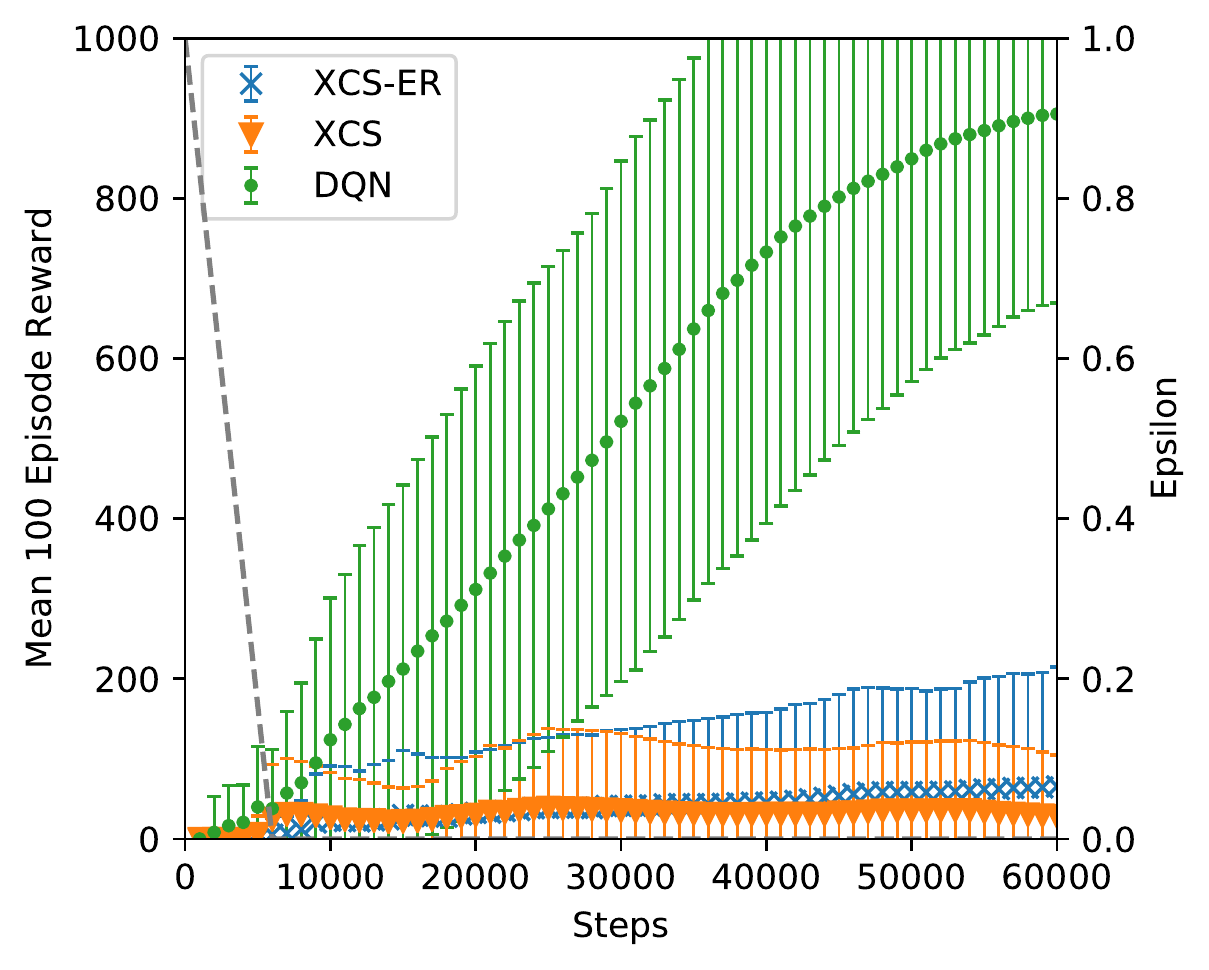}
    \caption{XCS vs. XCS-ER in MountainCar.}
\end{subfigure}
\qquad
\begin{subfigure}[b]{.47\textwidth}
    \centering
    \includegraphics[width=\textwidth]{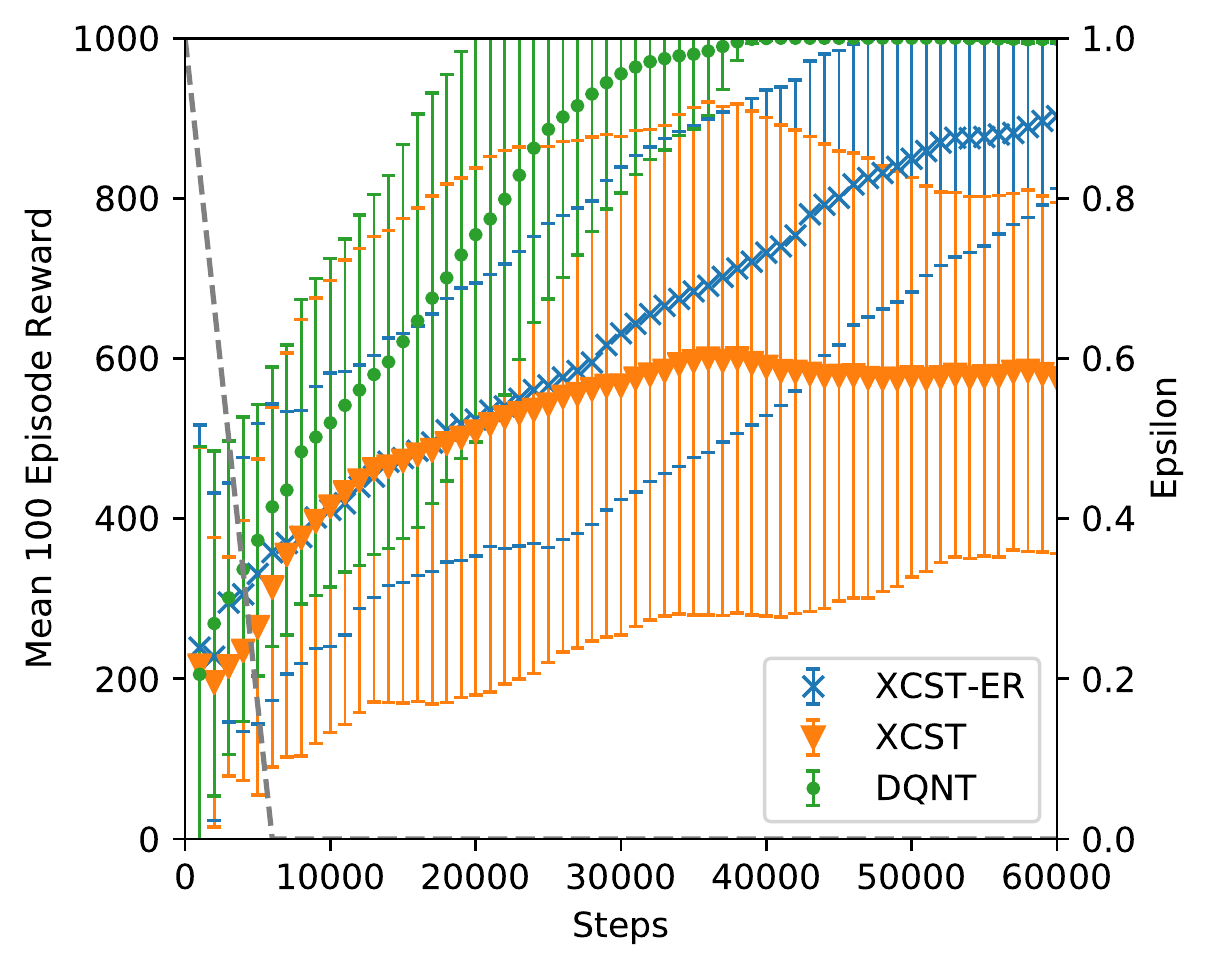}
    \caption{XCST vs. XCST-ER in MountainCar.}
\end{subfigure}  

\caption{XCS vs. XCS-ER vs. DQN on multi-step problems.}
\label{fig:eval_multistep}

\end{figure*}

The plots of the observed learning curves as shown in Figure~\ref{fig:eval_multistep} summarize the results. 
In order to provide a comparability with DQN, we show the \textit{overall total mean} (OTM) which calculates as the 100-episode mean of the received return (accumulated reward) per episode (as used in the OpenAI gym) averaged over the 30 experiment repetitions.
Additionally, in the supplemental material\footnote{We do not show the numerical results here, since for the following discussions the observed learning behavior at a higher level is descriptive enough to convey the relevant insights.} a similar result table containing the same metrics as for the single-step case is provided. There, also the number of \textit{divergences} observed across the repetitions is indicated in the last column. A divergence here is defined as an experimental run (until the max. number of steps is reached) where the algorithm was not able to find a proper solution and got stuck in local optima, i.e., never converged.

The results reveal that XCS with and without ER struggles in solving the examined multi-step environments (cf. the reported number of diverged runs). Especially in the MountainCar environment, for 29 out of 30 runs XCS was not able to solve the problem. In contrast, DQN only diverged 3 times. 
This observation was attributed to the well-known over-generalization issue of XCS when applied to certain multi-step scenarios. Due to the non-uniform sampling in sequential problems, classifiers quickly tend to overgeneralize across environmental niches~\cite{lanzi1999analysis,Barry2002}. This is because a prevalent lack in negative reward signals for situations that only barely or do not occur at all, which hinders XCS from identifying rules to be inaccurate in those regions.\footnote{Current research on a learning optimality theory~\cite{NakataBrowneHamagamiEtAl2017} is concerned with that issue.} This in turn blurs the fitness signal on which the employed GA bases its coverage or generality optimization. According to Wilson's \textit{generalization hypothesis}~\cite{wilson1995classifier}, XCS will generalize rules as long as the niche-relative fitness does not decrease. Without the ability to reliably identify so-called \textit{(strong) overgenerals}~\cite{Kovacs2000}, XCS continues to create even more general offspring rules which eventually swamp the population -- in the worst case leading to divergences.
In presence of a sparse reward signal, as artificially produced for the MountainCar environment in this paper, this issue becomes even more dramatic.
Exactly this phenomenon also constitutes a reason why XCS faces difficulties to sustain long-action-chains in multi-step environments~\cite{Barry2002}.

Our initial hypothesis states that we suspected that the application of ER would increase XCS's learning capability and thus attenuate this effect. Actually, quite the contrary appeared to be true.
ER increases sample efficiency by repeatedly feeding already made experiences into the credit assignment procedures (in case of XCS, the Widrow-Hoff update rule).
With an already skewed underlying experience or sample distribution, the minibatches drawn uniformly from the RM share the same skewness. 
Breaking the correlations between consecutively encountered experiences alone does not result in a more uniform prior data distribution regarding the samples from the state-action-reward landscape $S\times A \rightarrow \mathbb{R}$. 
In theory, XCS should not suffer from consecutive samples to the same extent as global approximators such as DQN do, since it already breaks the overall problem space up into several environmental niches for which an individual model (collectively determined by the residing classifiers) is sought each.
However, in practice it turns out that XCS is not yet at this point, which clearly requests for further research in this regard. 
To corroborate our abovementioned rationale, we modified the initial states of the environments to be more randomly distributed among the state space $S$. Therefore, we employed the \textit{teletransportation} technique~\cite{lanzi1999analysis}.
As can be seen from the corresponding plots showing the learning progress (see Fig.~\ref{fig:eval_multistep}), effectively changing the prior state distribution through teletransportation completely eradicates the occured divergences for XCS in the nChain and MountainCar scenario. 
Naturally, also DQN benefits from this treatment.  
Considering the CartPole environment, a more uniform distribution of the initial states dramatically increases the problem complexity. Now, a controller is sought which can stabilize the pole from any position, at any velocity and any starting angle and velocity of the pole. This demands for a much longer learning period, which is not subject of the present work. However, overall XCS-ER appears to still benefit from ER considering the OTM metric compared to standard XCS.    

The observed results for the multi-step case reveal that ER amplifies a well-recognized issue of XCS when applied to sequential control tasks demanding for sustenance of long-action-chains -- the occurrence of overgeneral classifiers bearing the risk of population divergence. 
Contrary to our initially posed hypothesis, ER aggravates this phenomenon by promoting misleading fitness signals through quicker convergence of the classifiers' error estimates in spite of actually being overly general. 
Paired with XCS's current deficiency in reliably identifying such actually inaccurate rules in turn fuels the employed GA which is responsible for seeking maximally general rules without loosing fitness (i.e., predictive accuracy).
This problem becomes even more dramatic in problems with continuous state spaces, demanding for hyperrectangular condition representations. Due to the adapted mutation operations, a lack of pressure toward more specified rules exists in contrast to ternary coded classifier conditions~\cite{Butz2005c}. 

Interestingly, throughout all experiments, DQN showed superior or at least similar performance compared to XCS with and without ER. 
So far, we did not fully understand the concrete reasons for that observation which prompts for further research in this direction. 

\section{Conclusion}\label{sec:conclusion}
We presented XCS-ER, an extended version of XCS which makes use of past experiences through an experience replay mechanism.
A literature review revealed that the interest in combining LCS with memory mechanisms providing access to raw experiences not further fed into an internal model building process has been rather low. 
The reported results of conducted empirical studies promise that ER leads to significant improvements for single-step problems. In contrast, after setting multi-step problems in the spotlight, it turned out that ER bears the risk to dramatically aggravate already recognized issues inherent to XCS when confronted with learning tasks demanding for establishment of long-action-chains; a domain where RL-approaches such as DQN prove better suited so far. 

Further research will be explicitly targeted at demystifying the not yet fully understood reasons for why DQN works so much better on difficult, i.e., long-action-chain demanding sequential control tasks such as MountainCar (see discussion above).
Deeper investigations of XCS regarding its still prevalent overgeneralization issue in long-action-chain environments have to be conducted, comprising the combination of ER with countermeasures against overgeneralization such as the recently proposed \emph{absumption} routine or the \emph{specify} operator as introduced by Liu et al.~\cite{Liu2019} and Lanzi~\cite{lanzi1999analysis}, respectively.
As soon as the identified pitfalls of integrating ER in XCS for multi-step problems are understood more clearly, more sophisticated ER strategies such as \textit{Hindsight ER}~\cite{andrychowicz2017hindsight}, \textit{Prioritized ER}~\cite{schaul2015prioritized}, or a recently proposed interpolation-based technique for bootstrapping the replay memory~\cite{pilchau2020bootstrapping} will serve as subject for further improvements.   

\bibliographystyle{ACM-Reference-Format}
\bibliography{bibliography}
\clearpage
\appendix

\vfill 
\section{Supplemental Material}
\subsection{Experiment Configurations}
\paragraph{Environment Modifications}

For the \textbf{n-Chain} environment, the episode length was reduced from 1000 to only 200, what increases the problem difficulty. 

For the \textbf{CartPole} environment, the input parameters have been re-scaled to the interval [0,1] (using reasonable lower and upper bounds as determined in preliminary runs) since the utilized XCS implementation expects normalized inputs. The reward scheme was slightly modified to deliver +1 every time the pole keeps balanced by the cart and 0 if it has fallen. This provides a clearer reward signal and facilitates quicker learning for each of the compared algorithms. 

Regarding the \textbf{MountainCar} environment, we again changed the episode length from 200 to 500, since both XCS variants encountered problems in randomly stumbling into the goal state within only 200 steps. Further, the reward scheme was made binary, i.e., instead of paying -1 for every step (includes reaching the terminal/goal state), we chose to apply a 0/1000 scheme. The latter only pays a reward of 1000 when reaching the goal, otherwise 0. This modification essentially increases the problem difficulty, since it renders the reward signal sparse.
\vfill 

\clearpage

\vfill

\begin{table*}[htbp]
\caption{Configured hyperparameter settings. Abbreviations used: TS = tournament selection, UBR = unordered bound hyperrectangular representation, WH = Widrow-Hoff learning rule, RLS = recursive least square learning rule}
\label{tab:hyperparameters}
\begin{tabular}{r c c c c c c c c c}
Parameter      & Mario & 6-RMP & WBC & 16chain & CartPole & M.Car &&& \\
\toprule
Learning steps   & 100k & 40k & 50k & 50k & 50k & 60k &&& \\
$N_{max}$   & 7000 & 800 & 6400 & 1000 & 5000 & 3000 &&& \\
$\beta$     & 0.3 & 0.2 & 0.2 & 0.1 & 0.1 & 0.1 &&& \\
$\gamma$    & - & - & - & 0.9 & 0.99 & 0.99 &&& \\
\textbf{Fitness parameters:} &&&&&&&&& \\
$\alpha$    & 0.1 & 0.1 & 0.1 & 0.1 & 0.1 & 0.1 &&& \\
$\epsilon_0$ & 10 & 10 & 1 & 0.1 & 8.0 & 0.1 &&& \\
$\nu$       & 5 & 5 & 5 & 5 & 5 & 5 &&& \\
\textbf{Deletion parameters:}   &&&&&&&&& \\
$\theta_{del}$ & 50 & 20 & 50 & 20 & 200 & 200 &&& \\
$\delta$    & 0.1 & 0.1 & 0.1 & 0.1 & 0.1 & 0.1 &&& \\
\textbf{Covering parameters:}   &&&&&&&&& \\
$\theta_{mna}$ & 7 & 2 & 2 & 2 & 2 & 3 &&& \\
$p_{ini}$   & 10 & 10 & 10 & 10 & 0.01 & 0.01 &&& \\
$\epsilon_{ini}$ & 0 & 0 & 0 & 10 & 0.01 & 0.01 &&& \\
$F_{ini}$   & 0.01 & 0.01 & 0.01 & 0.01 & 0.01 & 0.01 &&& \\
\textbf{GA parameters:}   &&&&&&&&& \\
$\mu$       & 0.04 & 0.04 & 0.04 & 0.04 & 0.04 & 0.04 &&& \\
$\chi$      & 0.8 & 0.8 & 0.8 & 0.8 & 0.8 & 0.8 &&& \\
$\theta_{GA}$ & 30 & 12 & 48 & 50 & 25 & 50 &&& \\
$\theta_{sub}$ & 50 & 20 & 50 & 200 & 20 & 20 &&& \\
GA-selection & TS & TS & TS & TS & TS & TS &&& \\
$F_{reduce}$ & 0.1 & 0.1 & 0.1 & 0.1 & 0.1 & 0.1 &&& \\
$\epsilon_{reduce}$ & 1.0 & 1.0 & 1.0 & 0.25 & 0.25 & 0.25 &&& \\
\textbf{Condition parameters:}   &&&&&&&&& \\
Condition   & UBR & UBR & UBR & UBR & UBR & UBR &&& \\
$m_0$       & 0.1 & 0.1 & 0.2 & 0.1 & 0.1 & 0.1 &&& \\
$r_0$       & 0.1 & 1.0 & 0.4 & 0.2 & 0.2 & 0.33 &&& \\
\textbf{Prediction parameters:}   &&&&&&&&& \\
Prediction   & scalar & scalar & scalar & scalar & linear & linear &&& \\
Learning rule   & WH & WH & WH & WH & RLS & RLS &&& \\
$\delta_{RLS}$ & - & - & - & - & 1.0 & 500 &&& \\
$\lambda_{RLS}$   & - & - & - & - & 1.0 & 1.0 &&& \\
\bottomrule
\end{tabular}
\end{table*}

For all experiments a maximum capacity for the Replay Memory (RM) of $N=50k$ has been chosen. The mini-batch size was set to $m=4$. In all experiments, a warm-up phase to fill the RM was conducted during the initial $1000$ steps. No learning happened in this phase.  

\clearpage

\subsection{Additional Statistics}

\begin{figure*}[htpb]
\centering
  \includegraphics[scale=0.9]{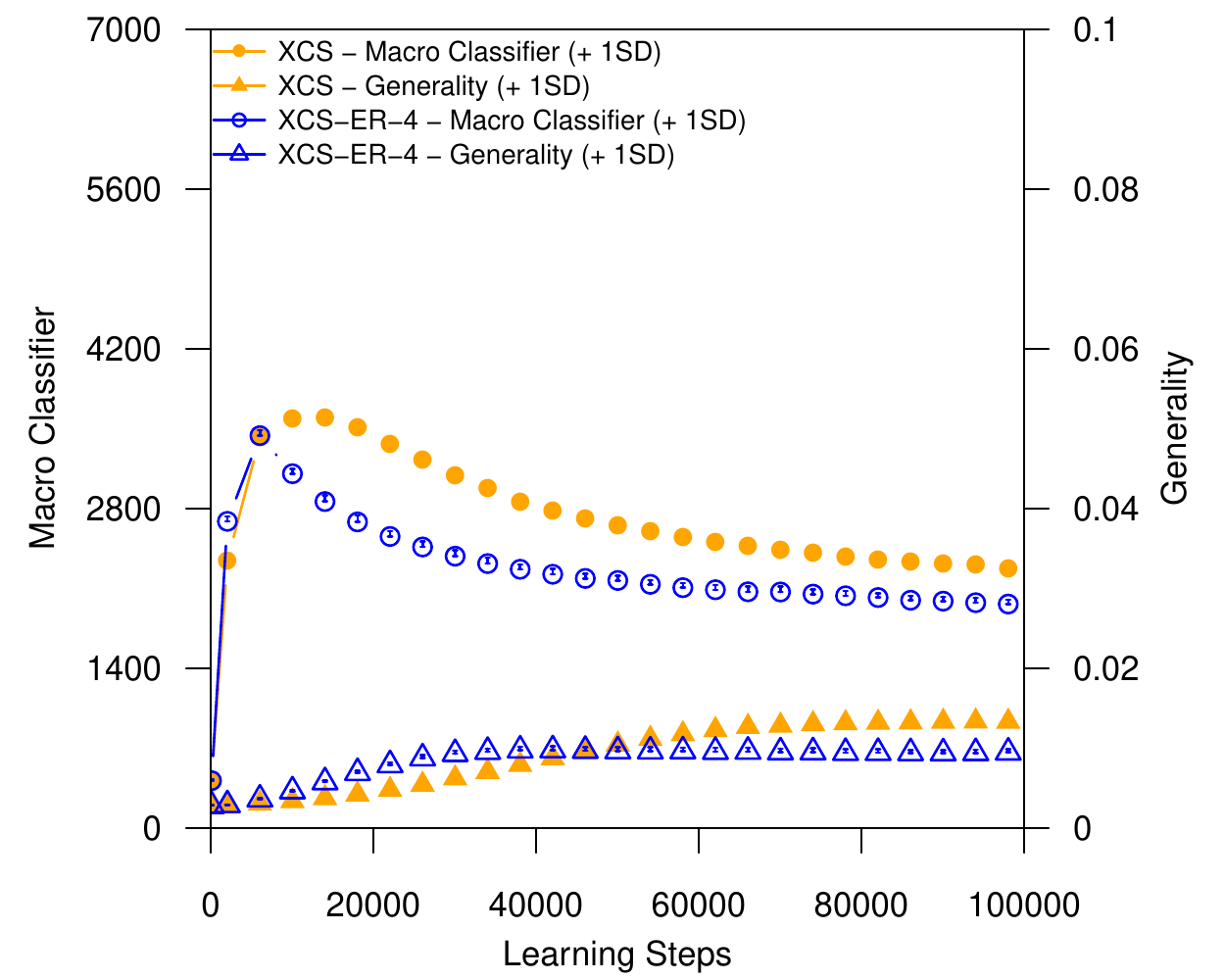}
    \caption{XCS vs. XCS-ER on Mario pixel art.}
\end{figure*}

\begin{table*}[htbp]
\caption{Shapiro-Wilk tests for normality on the Mario problem. The resulting p-values are given in the cells. * (**) indicates a p-value < $\alpha$ = 0.05 (0.01), i.e. a high likelihood that the null hypothesis (normality) can be rejected. p-values below 0.05 are marked in bold.}

\begin{tabular}{l|l|l|l|}
\cline{2-4}
\textbf{Mario} & XCS-ER-8 & XCS-ER-4 & XCS \\ \hline
\multicolumn{1}{|l|}{Reward mean} & 0.783 & 0.637 & 0.936 \\ \hline
\multicolumn{1}{|l|}{System error mean} & 0.467 & 0.858 & 0.065 \\ \hline
\multicolumn{1}{|l|}{Macroclassifiers mean} & 0.209 & 0.863 & 0.235 \\ \hline
\multicolumn{1}{|l|}{Generality mean} & 0.213 & 0.586 & 0.110 \\ \hline
\end{tabular}
\end{table*}

\begin{table*}[htbp]
\caption{Paired one-sided t-tests for XCS vs. XCS-ER on the Mario problem. The resulting p-values are given in the cells. * (**) indicates a p-value < $\alpha$ = 0.05 (0.01), i.e. a high likelihood the deviation is statistically significant. p-values below 0.05 are marked in bold.}

\begin{tabular}{l|l|l|}
\cline{2-3}
\textbf{Mario} & XCS-ER-8 & XCS-ER-4 \\ \hline
\multicolumn{1}{|l|}{Reward mean} & \textbf{0.000**} & \textbf{0.000**} \\ \hline
\multicolumn{1}{|l|}{System error mean} & \textbf{0.000**} & \textbf{0.000**} \\ \hline
\multicolumn{1}{|l|}{Macroclassifiers mean} & \textbf{0.000**} & \textbf{0.000**} \\ \hline
\multicolumn{1}{|l|}{Generality mean} & \textbf{0.000**} & \textbf{0.000**} \\ \hline
\end{tabular}
\end{table*}

\pagebreak

\begin{figure*}[htpb]
  \includegraphics[scale=0.9]{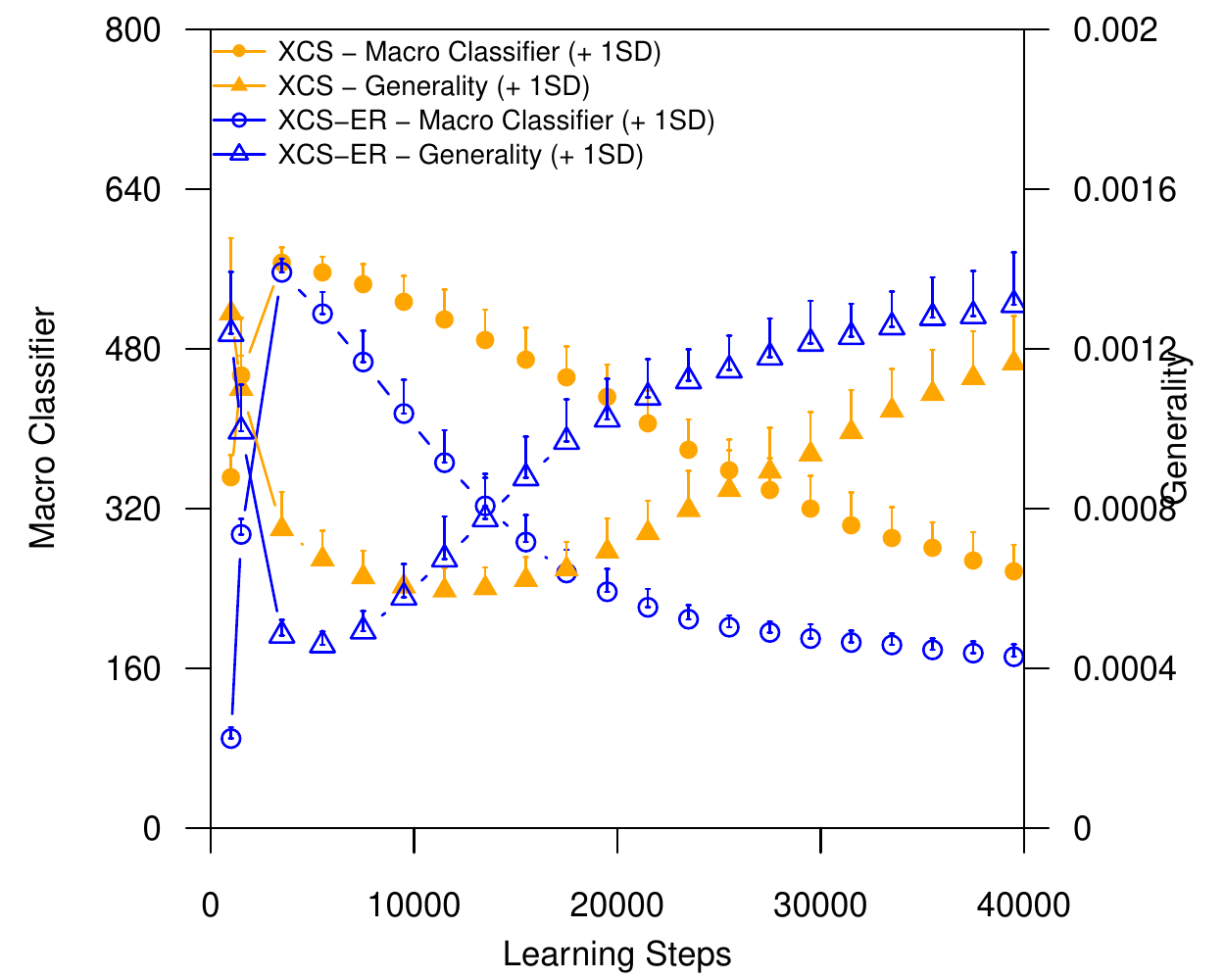}
    \caption{XCS vs. XCS-ER on 6-Real Multiplexer}
\end{figure*}

\begin{table*}[htbp]
\caption{Shapiro-Wilk tests for normality for the 6-RMP problem. The resulting p-values are given in the cells. * (**) indicates a p-value < $\alpha$ = 0.05 (0.01), i.e. a high likelihood that the null hypothesis (normality) can be rejected. p-values below 0.05 are marked in bold.}

\begin{tabular}{l|l|l|}
\cline{2-3}
\textbf{6-Real Multiplexer} & XCS-ER & XCS \\ \hline
\multicolumn{1}{|l|}{Reward mean} & 0.060 & 0.872 \\ \hline
\multicolumn{1}{|l|}{System error mean} & 0.143 & 0.714 \\ \hline
\multicolumn{1}{|l|}{Macroclassifiers mean} & \textbf{0.037*} & 0.882 \\ \hline
\multicolumn{1}{|l|}{Generality mean} & 0.846 & 0.117 \\ \hline
\end{tabular}
\end{table*}

\begin{table*}[htbp]
\caption{Paired one-sided t-tests for XCS vs. XCS-ER on the 6-RMP problem. The resulting p-values are given in the cells. * (**) indicates a p-value < $\alpha$ = 0.05 (0.01), i.e. a high likelihood the deviation from the baseline (XCS without ER) is statistically significant. p-values below 0.05 are marked in bold.}

\begin{tabular}{l|l|}
\cline{2-2}
\textbf{6-Real Multiplexer} & XCS-ER \\ \hline
\multicolumn{1}{|l|}{Reward mean} & \textbf{0.000**} \\ \hline
\multicolumn{1}{|l|}{System error mean} & \textbf{0.000**} \\ \hline
\multicolumn{1}{|l|}{Macroclassifiers mean} & \textbf{0.000**} \\ \hline
\multicolumn{1}{|l|}{Generality mean} & \textbf{0.000**} \\ \hline
\end{tabular}
\end{table*}

\pagebreak

\begin{figure*}[htbp]
  \includegraphics[scale = 0.9]{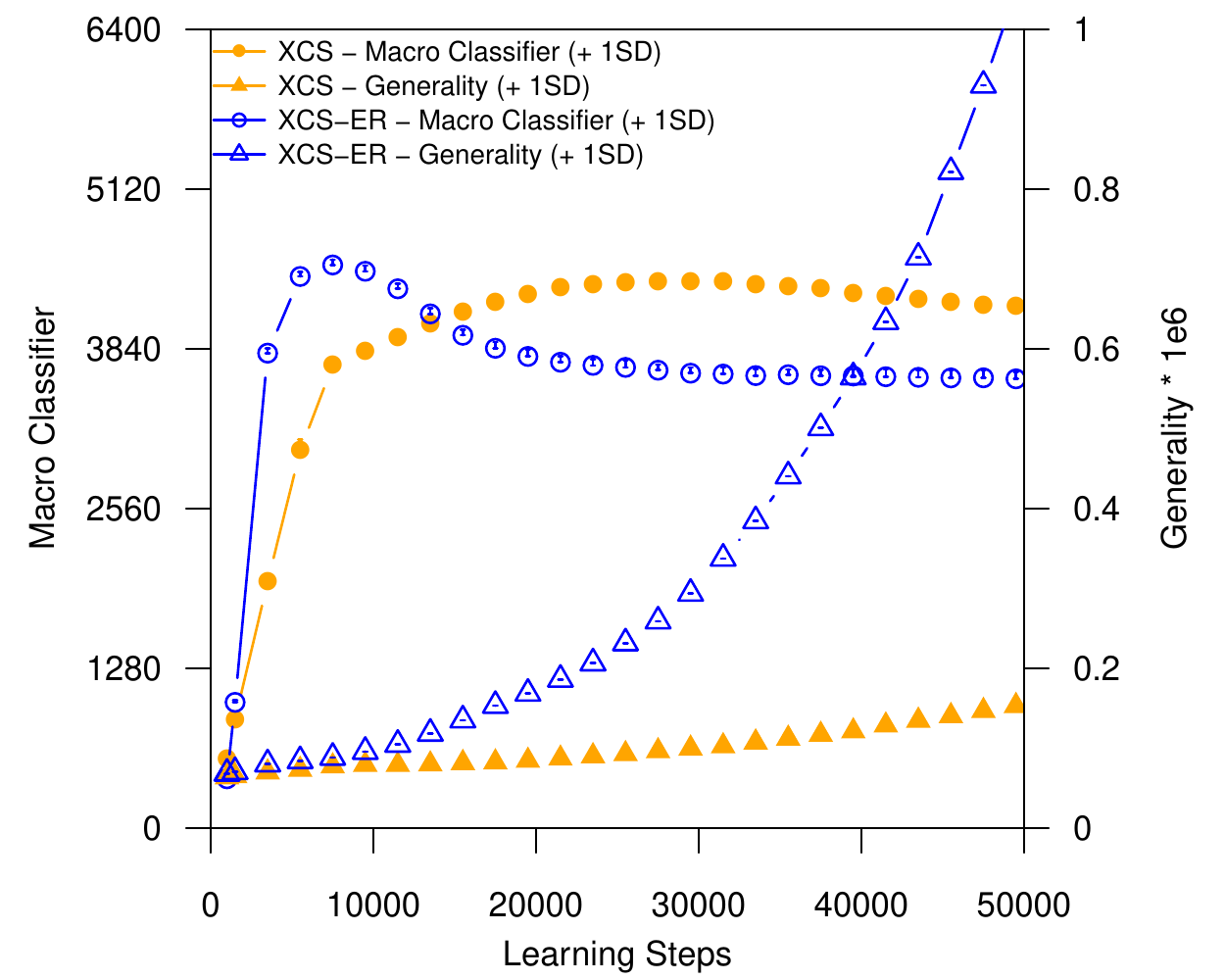}
  \caption{XCS vs. XCS-ER on WBC classification}

\end{figure*}  

\begin{table*}[htpb]
\caption{Shapiro-Wilk tests for normality on the WBC data set. The resulting p-values are depicted in the cells. * (**) indicates a p-value < $\alpha$ = 0.05 (0.01), i.e. a high likelihood that the null hypothesis (normality) can be rejected. p-values below 0.05 are marked in bold.}

\begin{tabular}{l|l|l|}
\cline{2-3}
\textbf{WBC} & XCS-ER & XCS \\ \hline
\multicolumn{1}{|l|}{Reward mean} & 0.458 & 0.333 \\ \hline
\multicolumn{1}{|l|}{System error mean} & 0.195 & 0.316 \\ \hline
\multicolumn{1}{|l|}{Macroclassifiers mean} & 0.096 & 0.458 \\ \hline
\multicolumn{1}{|l|}{Generality mean} & \textbf{0.026*} & \textbf{0.003**} \\ \hline
\end{tabular}
\end{table*}

\begin{table*}[htpb]
\caption{Paired one-sided t-tests for XCS vs. XCS-ER on the WBC data set. The resulting p-values are given in the cells. * (**) indicates a p-value < $\alpha$ = 0.05 (0.01), i.e. a high likelihood the deviation from the baseline (XCS without ER) is statistically significant. p-values below 0.05 are marked in bold.}

\begin{tabular}{l|l|}
\cline{2-2}
\textbf{WBC} & XCS-ER \\ \hline
\multicolumn{1}{|l|}{Reward mean} & \textbf{0.000**} \\ \hline
\multicolumn{1}{|l|}{System error mean} & \textbf{0.000**} \\ \hline
\multicolumn{1}{|l|}{Macroclassifiers mean} & \textbf{0.000**} \\ \hline
\multicolumn{1}{|l|}{Generality mean} & \textbf{0.000**} \\ \hline
\end{tabular}
\end{table*}

\begin{table*}[htpb]
\caption{Overall results for the multi-step environments: Reward, prediction error, population size in macroclassifiers, generality, OTM and Divergences are shown. Table entries indicate the overall means $\pm$1SD over the 30 repetitions. Divergence criteria for individual repetitions for 16Chain, CartPole, MountainCar are OTM values below 1, 70 and 200 respectively. 'T' indicates active teletransportation. Arrows show whether the metric increased up or decreased. Statistical significance when compared to the regular XCS is determined via the Wilcoxon signed-rank test. * (**) indicates a p-value < $\alpha$ = 0.05 (0.01). p-values below 0.05 confirming improvements are given in bold.}
\label{tab:multi-step}
\centering
\begin{tabular}{rllllll}
\bottomrule
\multicolumn{1}{c}{\textbf{16Chain}} & 
\multicolumn{1}{l}{\begin{tabular}[c]{@{}c@{}}Reward\end{tabular}} & \multicolumn{1}{l}{\begin{tabular}[c]{@{}c@{}}Pr.Err.\end{tabular}} & \multicolumn{1}{l}{\begin{tabular}[c]{@{}c@{}}$|[P]|$\end{tabular}} & \multicolumn{1}{l}{\begin{tabular}[c]{@{}c@{}}Gen. \end{tabular}} & 
\multicolumn{1}{l}{OTM} & 
\multicolumn{1}{l}{\begin{tabular}[c]{@{}c@{}}Div\end{tabular}} \\ 
\toprule
\multicolumn{1}{r}{XCS-ER} & \begin{tabular}[c]{@{}l@{}}0.56$\downarrow$**\\ $\pm$0.37\end{tabular} & \begin{tabular}[c]{@{}l@{}}0.23$\uparrow$**\\ $\pm$0.01\end{tabular} & \textbf{\begin{tabular}[c]{@{}l@{}}92.92$\downarrow$**\\ $\pm$25.81\end{tabular}} & \textbf{\begin{tabular}[c]{@{}l@{}}0.13$\uparrow$**\\ $\pm$0.06\end{tabular}} & \begin{tabular}[c]{@{}l@{}}95.3$\downarrow$**\\ $\pm$63.6\end{tabular} & 9$\uparrow$\\ 
\hline
\multicolumn{1}{r}{DQN} & n/a & n/a & n/a & n/a & \begin{tabular}[c]{@{}l@{}}143.1$\uparrow$\\ $\pm$1.4\end{tabular} & 0$\downarrow$ \\ 
\hhline{=======}
\multicolumn{1}{r}{XCS} & \begin{tabular}[c]{@{}l@{}}0.82\\ $\pm$0.15\end{tabular} & \begin{tabular}[c]{@{}l@{}}0.14\\ $\pm$0.03\end{tabular} & \begin{tabular}[c]{@{}l@{}}152.68\\ $\pm$22.16\end{tabular} & \begin{tabular}[c]{@{}l@{}}0.07\\ $\pm$0.02\end{tabular} & \begin{tabular}[c]{@{}l@{}}138.6\\ $\pm$25.7\end{tabular} & 1 \\ 
\bottomrule
\multicolumn{1}{c}{\textbf{16Chain T}} & 
\multicolumn{1}{l}{\begin{tabular}[c]{@{}c@{}}Reward\end{tabular}} & \multicolumn{1}{l}{\begin{tabular}[c]{@{}c@{}}Pr.Err.\end{tabular}} & \multicolumn{1}{l}{\begin{tabular}[c]{@{}c@{}}$|[P]|$\end{tabular}} & \multicolumn{1}{l}{\begin{tabular}[c]{@{}c@{}}Gen. \end{tabular}} & 
\multicolumn{1}{l}{OTM} & 
\multicolumn{1}{l}{\begin{tabular}[c]{@{}c@{}}Div\end{tabular}} \\ 
\toprule\multicolumn{1}{r}{XCST-ER} & \begin{tabular}[c]{@{}l@{}}0.86$\downarrow$\\ $\pm$0.03\end{tabular} & \begin{tabular}[c]{@{}l@{}}0.24$\uparrow$**\\ $\pm$0.01\end{tabular} & \begin{tabular}[c]{@{}l@{}}111.65$\downarrow$\\ $\pm$21.94\end{tabular} & \begin{tabular}[c]{@{}l@{}}0.10$\uparrow$\\ $\pm$0.05\end{tabular} & \begin{tabular}[c]{@{}l@{}}147.6$\downarrow$\\ $\pm$4.2\end{tabular} & 0 \\ 
\hline
\multicolumn{1}{r}{DQNT} & n/a & n/a & n/a & n/a & \begin{tabular}[c]{@{}l@{}}148.0$\downarrow$\\ $\pm$5.8\end{tabular} & 0 \\ 
\hhline{=======}
\multicolumn{1}{r}{XCST} & \begin{tabular}[c]{@{}l@{}}0.88\\ $\pm$0.00\end{tabular} & \begin{tabular}[c]{@{}l@{}}0.20\\ $\pm$0.03\end{tabular} & \begin{tabular}[c]{@{}l@{}}117.90\\ $\pm$13.54\end{tabular} & \begin{tabular}[c]{@{}l@{}}0.09\\ $\pm$0.01\end{tabular} & \begin{tabular}[c]{@{}l@{}}149.2\\ $\pm$0.6\end{tabular} & 0 \\ 
\bottomrule
\multicolumn{1}{c}{\textbf{CartPole}} & 
\multicolumn{1}{l}{\begin{tabular}[c]{@{}c@{}}Reward\end{tabular}} & \multicolumn{1}{l}{\begin{tabular}[c]{@{}c@{}}Pr.Err.\end{tabular}} & \multicolumn{1}{l}{\begin{tabular}[c]{@{}c@{}}$|[P]|$\end{tabular}} & \multicolumn{1}{l}{\begin{tabular}[c]{@{}c@{}}Gen. \end{tabular}} & 
\multicolumn{1}{l}{OTM} & 
\multicolumn{1}{l}{\begin{tabular}[c]{@{}c@{}}Div\end{tabular}} \\ 
\toprule
\multicolumn{1}{r}{XCS-ER} & \begin{tabular}[c]{@{}l@{}}0.98$\downarrow$\\ $\pm$0.03\end{tabular} & \begin{tabular}[c]{@{}l@{}}1.92$\uparrow$**\\ $\pm$0.11\end{tabular} & \begin{tabular}[c]{@{}l@{}}1937.44$\uparrow$**\\ $\pm$106.85\end{tabular} & \textbf{\begin{tabular}[c]{@{}l@{}}0.0004$\uparrow$**\\ $\pm$0.0000\end{tabular}} & \begin{tabular}[c]{@{}l@{}}93.0$\downarrow$\\ $\pm$47.6\end{tabular} & 9$\uparrow$ \\ \hline
\multicolumn{1}{r}{DQN} & n/a & n/a & n/a & n/a & \begin{tabular}[c]{@{}l@{}}122.2$\uparrow$\\ $\pm$13.4\end{tabular} & 0$\downarrow$ \\ 
\hhline{=======}
\multicolumn{1}{r}{XCS} & \begin{tabular}[c]{@{}l@{}}0.99\\ $\pm$0.01\end{tabular} & \begin{tabular}[c]{@{}l@{}}1.62\\ $\pm$0.25\end{tabular} & \begin{tabular}[c]{@{}l@{}}1703.42\\ $\pm$161.46\end{tabular} & \begin{tabular}[c]{@{}l@{}}0.0003\\ $\pm$0.0000\end{tabular} & \begin{tabular}[c]{@{}l@{}}110.6\\ $\pm$46.5\end{tabular} & 7 \\ 
\bottomrule
\multicolumn{1}{c}{\textbf{CartPole T}} & 
\multicolumn{1}{l}{\begin{tabular}[c]{@{}c@{}}Reward\end{tabular}} & \multicolumn{1}{l}{\begin{tabular}[c]{@{}c@{}}Pr.Err.\end{tabular}} & \multicolumn{1}{l}{\begin{tabular}[c]{@{}c@{}}$|[P]|$\end{tabular}} & \multicolumn{1}{l}{\begin{tabular}[c]{@{}c@{}}Gen. \end{tabular}} & 
\multicolumn{1}{l}{OTM} & 
\multicolumn{1}{l}{\begin{tabular}[c]{@{}c@{}}Div\end{tabular}} \\ 
\toprule\multicolumn{1}{r}{XCST-ER} & \textbf{\begin{tabular}[c]{@{}l@{}}0.94$\uparrow$**\\ $\pm$0.02\end{tabular}} & \begin{tabular}[c]{@{}l@{}}2.17$\uparrow$**\\ $\pm$0.12\end{tabular} & \textbf{\begin{tabular}[c]{@{}l@{}}2162.30$\downarrow$**\\ $\pm$51.94\end{tabular}} & \textbf{\begin{tabular}[c]{@{}l@{}}0.0013$\uparrow$**\\ $\pm$0.0000\end{tabular}} & \textbf{\begin{tabular}[c]{@{}l@{}}18.29$\uparrow$**\\ $\pm$6.22\end{tabular}} & 30 \\ 
\hline
\multicolumn{1}{r}{DQNT} & n/a & n/a & n/a & n/a & \textbf{\begin{tabular}[c]{@{}l@{}}38.75**\\ $\pm$2.19\end{tabular}} & 30 \\ 
\hhline{=======}
\multicolumn{1}{r}{XCST} & \begin{tabular}[c]{@{}l@{}}0.92\\ $\pm$0.02\end{tabular} & \begin{tabular}[c]{@{}l@{}}1.90\\ $\pm$0.09\end{tabular} & \begin{tabular}[c]{@{}l@{}}2640.10\\ $\pm$70.37\end{tabular} & \begin{tabular}[c]{@{}l@{}}0.0005\\ $\pm$0.0001\end{tabular} & \begin{tabular}[c]{@{}l@{}}13.76\\ $\pm$3.59\end{tabular} & 30 \\ 
\bottomrule
\multicolumn{1}{c}{\textbf{M.Car}} & 
\multicolumn{1}{l}{\begin{tabular}[c]{@{}c@{}}Reward\end{tabular}} & \multicolumn{1}{l}{\begin{tabular}[c]{@{}c@{}}Pr.Err.\end{tabular}} & \multicolumn{1}{l}{\begin{tabular}[c]{@{}c@{}}$|[P]|$\end{tabular}} & \multicolumn{1}{l}{\begin{tabular}[c]{@{}c@{}}Gen. \end{tabular}} & 
\multicolumn{1}{l}{OTM} & 
\multicolumn{1}{l}{\begin{tabular}[c]{@{}c@{}}Div\end{tabular}} \\ 
\toprule
\multicolumn{1}{r}{XCS-ER} & \begin{tabular}[c]{@{}l@{}}0.13$\uparrow$\\ $\pm$0.06\end{tabular} & \textbf{\begin{tabular}[c]{@{}l@{}}1.73$\downarrow$**\\ $\pm$2.77\end{tabular}} & \begin{tabular}[c]{@{}l@{}}383.25$\uparrow$\\ $\pm$204.68\end{tabular} & \textbf{\begin{tabular}[c]{@{}l@{}}0.05$\uparrow$**\\ $\pm$0.01\end{tabular}} & \begin{tabular}[c]{@{}l@{}}38.57$\uparrow$\\ $\pm$86.37\end{tabular} & 29 \\ 
\hline
\multicolumn{1}{r}{DQN} & n/a & n/a & n/a & n/a & \textbf{\begin{tabular}[c]{@{}l@{}}504.96$\uparrow$**\\ $\pm$216.03\end{tabular}} & 3$\downarrow$ \\ 
\hhline{=======}
\multicolumn{1}{r}{XCS} & \begin{tabular}[c]{@{}l@{}}0.06\\ $\pm$0.15\end{tabular} & \begin{tabular}[c]{@{}l@{}}3.5E5\\ $\pm$1.8E6\end{tabular} & \begin{tabular}[c]{@{}l@{}}325.17\\ $\pm$162.39\end{tabular} & \begin{tabular}[c]{@{}l@{}}0.04\\ $\pm$0.01\end{tabular} & \begin{tabular}[c]{@{}l@{}}29.15\\ $\pm$63.27\end{tabular} & 29 \\ 
\bottomrule
\multicolumn{1}{c}{\textbf{M.Car T}} & 
\multicolumn{1}{l}{\begin{tabular}[c]{@{}c@{}}Reward\end{tabular}} & \multicolumn{1}{l}{\begin{tabular}[c]{@{}c@{}}Pr.Err.\end{tabular}} & \multicolumn{1}{l}{\begin{tabular}[c]{@{}c@{}}$|[P]|$\end{tabular}} & \multicolumn{1}{l}{\begin{tabular}[c]{@{}c@{}}Gen. \end{tabular}} & 
\multicolumn{1}{l}{OTM} & 
\multicolumn{1}{l}{\begin{tabular}[c]{@{}c@{}}Div\end{tabular}} \\ 
\toprule\multicolumn{1}{r}{XCST-ER} & \textbf{\begin{tabular}[c]{@{}l@{}}4.33$\uparrow$*\\ $\pm$1.3\end{tabular}} & \textbf{\begin{tabular}[c]{@{}l@{}}12.40$\downarrow$**\\ $\pm$3.69\end{tabular}} & \begin{tabular}[c]{@{}l@{}}1150.36$\uparrow$\\ $\pm$88.24\end{tabular} & \begin{tabular}[c]{@{}l@{}}0.01$\downarrow$**\\ $\pm$0.00\end{tabular} & \begin{tabular}[c]{@{}l@{}}626.3$\uparrow$\\ $\pm$132.4\end{tabular} & 0$\downarrow$ \\ 
\hline
\multicolumn{1}{r}{DQNT} & n/a & n/a & n/a & n/a & \textbf{\begin{tabular}[c]{@{}l@{}}814.5$\uparrow$**\\ $\pm$90.9\end{tabular}} & 0$\downarrow$ \\ 
\hhline{=======}
\multicolumn{1}{r}{XCST} & \begin{tabular}[c]{@{}l@{}}3.22\\ $\pm$2.17\end{tabular} & \begin{tabular}[c]{@{}l@{}}1.2E21\\ $\pm$6.3E21\end{tabular} & \begin{tabular}[c]{@{}l@{}}1041.51\\ $\pm$114.82\end{tabular} & \begin{tabular}[c]{@{}l@{}}0.02\\ $\pm$0.00\end{tabular} & \begin{tabular}[c]{@{}l@{}}511.2\\ $\pm$219.5\end{tabular} & 5 \\ 
\end{tabular}
\end{table*}

\clearpage

\begin{table*}[htpb]
\caption{Shapiro-Wilk tests for normality on the multi-step environments. The resulting p-values are given in the cells. * (**) indicates a p-value < $\alpha$ = 0.05 (0.01), i.e. a high likelihood that the null hypothesis (normality) can be rejected. p-values below 0.05 are marked in bold.}

\begin{tabular}{l|l|l|l|}
\cline{2-4}
\textbf{16Chain} & XCS-ER & XCS & DQN \\ \hline
\multicolumn{1}{|l|}{Reward mean} & \textbf{0.000**} & \textbf{0.000**} & n/a \\ \hline
\multicolumn{1}{|l|}{Prediction error mean} & \textbf{0.003**} & 0.366 & n/a \\ \hline
\multicolumn{1}{|l|}{Macroclassifiers mean} & 0.058 & \textbf{0.008**} & n/a \\ \hline
\multicolumn{1}{|l|}{Generality mean} & \textbf{0.000**} & \textbf{0.000**} & n/a \\ \hline
\multicolumn{1}{|l|}{OTM} & \textbf{0.000**} & \textbf{0.000**} & \textbf{0.000**} \\ \hline
\textbf{16Chain Tele.} & XCS-ER & XCS & DQN \\ \hline
\multicolumn{1}{|l|}{Reward mean} & \textbf{0.000**} & \textbf{0.037*} & n/a \\ \hline
\multicolumn{1}{|l|}{Prediction error mean} & \textbf{0.001**} & 0.531 & n/a \\ \hline
\multicolumn{1}{|l|}{Macroclassifiers mean} & \textbf{0.000**} & 0.597 & n/a \\ \hline
\multicolumn{1}{|l|}{Generality mean} & \textbf{0.000**} & 0.733 & n/a \\ \hline
\multicolumn{1}{|l|}{OTM} & \textbf{0.000**} & \textbf{0.02*} & \textbf{0.000**} \\ \hline
\textbf{CartPole} & XCS-ER & XCS & DQN \\ \hline
\multicolumn{1}{|l|}{Reward mean} & \textbf{0.000**} & \textbf{0.000**} & n/a \\ \hline
\multicolumn{1}{|l|}{Prediction error mean} & 0.625 & \textbf{0.023*} & n/a \\ \hline
\multicolumn{1}{|l|}{Macroclassifiers mean} & 0.875 & 0.222 & n/a \\ \hline
\multicolumn{1}{|l|}{Generality mean} & 0.432 & 0.618 & n/a \\ \hline
\multicolumn{1}{|l|}{OTM} & \textbf{0.006**} & \textbf{0.000**} & 0.099 \\ \hline
\textbf{CartPole Tele.} & XCS-ER & XCS & DQN \\ \hline
\multicolumn{1}{|l|}{Reward mean} & 0.056 & 0.639 & n/a \\ \hline
\multicolumn{1}{|l|}{Prediction error mean} & 0.184 & 0.167 & n/a \\ \hline
\multicolumn{1}{|l|}{Macroclassifiers mean} & \textbf{0.026*} & 0.976 & n/a \\ \hline
\multicolumn{1}{|l|}{Generality mean} & 0.594 & 0.240 & n/a \\ \hline
\multicolumn{1}{|l|}{OTM} & 0.177 & \textbf{0.002**} & 0.370 \\ \hline
\textbf{MountainCar} & XCS-ER & XCS & DQN \\ \hline
\multicolumn{1}{|l|}{Reward mean} & \textbf{0.000**} & \textbf{0.000**} & n/a \\ \hline
\multicolumn{1}{|l|}{Prediction error mean} & \textbf{0.000**} & \textbf{0.000**} & n/a \\ \hline
\multicolumn{1}{|l|}{Macroclassifiers mean} & \textbf{0.000**} & \textbf{0.006**} & n/a \\ \hline
\multicolumn{1}{|l|}{Generality mean} & 0.367 & 0.303 & n/a \\ \hline
\multicolumn{1}{|l|}{OTM} & \textbf{0.000**} & \textbf{0.000**} & \textbf{0.023*} \\ \hline
\textbf{MountainCar Tele.} & XCS-ER & XCS & DQN \\ \hline
\multicolumn{1}{|l|}{Reward mean} & 0.673 & 0.067 & n/a \\ \hline
\multicolumn{1}{|l|}{Prediction error mean} & 0.641 & \textbf{0.000**} & n/a \\ \hline
\multicolumn{1}{|l|}{Macroclassifiers mean} & \textbf{0.024*} & \textbf{0.023*} & n/a \\ \hline
\multicolumn{1}{|l|}{Generality mean} & 0.070 & \textbf{0.000**} & n/a \\ \hline
\multicolumn{1}{|l|}{OTM} & \textbf{0.004**} & 0.042 & \textbf{0.005**} \\ \hline
\end{tabular}
\end{table*}

As can be noticed, most of the metrics do not follow a normal distribution. Even if normality can be confirmed in certain cases, a parametric test is still not plausibly applicable when the sample considered for comparison violates this assumption. For that reason, we decided to exclusively conduct non-parametric Wilcoxon tests across all multi-step experiments.   

\begin{table*}[htpb]
\caption{Wilcoxon signed-rank tests for the multi-step environments. The resulting p-values are given in the cells. * (**) indicates a p-value < $\alpha$ = 0.05 (0.01), i.e. a high likelihood that the deviation from the baseline (XCS without ER) is statistically significant. p-values below 0.05 are marked in bold.}

\begin{tabular}{l|l|l|}
\cline{2-3}
\textbf{16Chain} & XCS-ER & DQN \\ \hline
\multicolumn{1}{|l|}{Reward mean} & \textbf{0.000**} & n/a \\ \hline
\multicolumn{1}{|l|}{Prediction error mean} & \textbf{0.000**} & n/a \\ \hline
\multicolumn{1}{|l|}{Macroclassifiers mean} & \textbf{0.000**} & n/a \\ \hline
\multicolumn{1}{|l|}{Generality mean} & \textbf{0.000**} & n/a \\ \hline
\multicolumn{1}{|l|}{OTM} & \textbf{0.000**} & 0.959 \\ \hline
\textbf{16Chain Tele.} & XCS-ER & DQN \\ \hline
\multicolumn{1}{|l|}{Reward mean} & 0.254 & n/a \\ \hline
\multicolumn{1}{|l|}{Prediction error mean} & \textbf{0.000**} & n/a \\ \hline
\multicolumn{1}{|l|}{Macroclassifiers mean} & 0.349 & n/a \\ \hline
\multicolumn{1}{|l|}{Generality mean} & 0.644 & n/a \\ \hline
\multicolumn{1}{|l|}{OTM} & 0.394 & 0.504 \\ \hline
\textbf{CartPole} & XCS-ER & DQN \\ \hline
\multicolumn{1}{|l|}{Reward mean} & 0.136 & n/a \\ \hline
\multicolumn{1}{|l|}{Prediction error mean} & \textbf{0.000**} & n/a \\ \hline
\multicolumn{1}{|l|}{Macroclassifiers mean} & \textbf{0.000**} & n/a \\ \hline
\multicolumn{1}{|l|}{Generality mean} & \textbf{0.000**} & n/a \\ \hline
\multicolumn{1}{|l|}{OTM} & 0.111 & 0.428 \\ \hline
\textbf{CartPole Tele.} & XCS-ER & DQN \\ \hline
\multicolumn{1}{|l|}{Reward mean} & \textbf{0.000**} & n/a \\ \hline
\multicolumn{1}{|l|}{Prediction error mean} & \textbf{0.000**} & n/a \\ \hline
\multicolumn{1}{|l|}{Macroclassifiers mean} & \textbf{0.000**} & n/a \\ \hline
\multicolumn{1}{|l|}{Generality mean} & \textbf{0.000**} & n/a \\ \hline
\multicolumn{1}{|l|}{OTM} & \textbf{0.000**} & \textbf{0.000**} \\ \hline
\textbf{MountainCar} & XCS-ER & DQN \\ \hline
\multicolumn{1}{|l|}{Reward mean} & 0.178 & n/a \\ \hline
\multicolumn{1}{|l|}{Prediction error mean} & \textbf{0.002**} & n/a \\ \hline
\multicolumn{1}{|l|}{Macroclassifiers mean} & 0.360 & n/a \\ \hline
\multicolumn{1}{|l|}{Generality mean} & \textbf{0.001**} & n/a \\ \hline
\multicolumn{1}{|l|}{OTM} & 0.550 & \textbf{0.000**} \\ \hline
\textbf{MountainCar Tele.} & XCS-ER & DQN \\ \hline
\multicolumn{1}{|l|}{Reward mean} & \textbf{0.014*} & n/a \\ \hline
\multicolumn{1}{|l|}{Prediction error mean} & \textbf{0.000**} & n/a \\ \hline
\multicolumn{1}{|l|}{Macroclassifiers mean} & \textbf{0.000**} & n/a \\ \hline
\multicolumn{1}{|l|}{Generality mean} & \textbf{0.000**} & n/a \\ \hline
\multicolumn{1}{|l|}{OTM} & 0.060 & \textbf{0.000**} \\ \hline
\end{tabular}
\end{table*}

\end{document}